\RequirePackage{fix-cm}
\documentclass[smallcondensed]{svjour3}     % onecolumn (ditto)
\smartqed  % flush right qed marks, e.g. at end of proof
\usepackage{graphicx}
\usepackage{color,soul}
\usepackage{changes}
\soulregister\ref{7}
\soulregister\cite{7}
\soulregister\citep{7}
\soulregister\item{7}
\usepackage{amsmath}
\usepackage{changes}
\usepackage{ifpdf}
\usepackage{textcomp}
\usepackage {amssymb}
\usepackage{subfig}
\usepackage{stmaryrd}
\usepackage{url}
\usepackage{listings}
\usepackage{enumitem}
\usepackage{amstext}
\usepackage{rotating}
\usepackage{changes}

\usepackage{array}
\usepackage{tikz}
\usetikzlibrary{arrows,automata}
\usepackage{qtree}
\usepackage{multirow}
\usetikzlibrary{positioning}
\definecolor{mybluei}{RGB}{124,156,205}
\definecolor{myblueii}{RGB}{73,121,193}
\definecolor{mygreen}{RGB}{202,217,126}
\usepackage{algorithm}
\usepackage{algorithmicx}

\usepackage{algpseudocode}

\newtheorem {onto}{Definition}
\newtheorem {Ex}{Example}

\begin{document}

\title{A Heuristically Modified FP-Tree for Ontology Learning with Applications in Education%
%Insert your title here%\thanks{Grants or other notes
}

\titlerunning{A Heuristically Modified FP-Tree for Ontology Learning}        % if too long for running head

\author{Safwan Shatnawi         \and
				Mohamed Medhat Gaber $^*$ 	\and
        Mihaela Cocea %etc.
}


\institute{S. Shatnawi \at
              College of Applied Studies, University of Bahrain, Sakhair Campus, Zallaq, Bahrain\\
              \email{smahmood@uob.edu.bh}           %  \\
           \and
					 M. M. Gaber \at
              School of Computing and Digital Technology, Birmingham City University, UK\\
							\email{mohamed.gaber@bcu.ac.uk}
					 \and
           M. Cocea \at
              School of computing  University of Portsmouth, UK\\
							\email{mihaela.cocea@port.ac.uk}
}

\date{Received: date / Accepted: date}

\maketitle

\begin{abstract}
We propose a heuristically modified FP-Tree for ontology learning from text. Unlike previous research, for concept extraction, we use a regular expression parser approach widely adopted in compiler construction, i.e., deterministic finite automata (DFA). Thus, the concepts are extracted from unstructured documents. For ontology learning, we use a frequent pattern mining approach and employ a rule mining heuristic function to enhance its quality. This process does not rely on predefined lexico-syntactic patterns, thus, it is applicable for different subjects. We employ the ontology in a question-answering system for students' content-related questions. For validation, we used textbook questions/answers and questions from online course forums. Subject experts rated the quality of the system's answers on a subset of questions and their ratings were used to identify the most appropriate automatic semantic text similarity metric to use as a validation metric for all answers. The Latent Semantic Analysis was identified as the closest to the experts' ratings. We compared the use of our ontology with the use of Text2Onto for the question-answering system and found that with our ontology 80\% of the questions were answered, while with Text2Onto only 28.4\% were answered, thanks to the finer grained hierarchy our approach is able to produce.

\end{abstract}

%**************************************************************************************************************************************************************************************************************************************************************************************************************************************************************
\section{Introduction}\label{sec:introduction}

Ontologies form the main knowledge structure of the semantic web. There is, however, a consensus among researchers that building and maintaining ontologies are expensive and time consuming tasks. 
In the learning technologies area most researchers either manually build a domain-specific ontology or assume the existence of such an ontology \cite{Instructor_Dunwei}, \cite{Measuring_Jianhua}. 

Ontologies have been used in the field of learning technology for various purposes such as instructional design~\cite{Asemantic_Seiji}, adaptive intelligent educational systems~\cite{Reasoning_Henze}, tutorial dialog systems~\cite{Automating_Fiedler}, assessment~\cite{Leveraging_Kazi}, feedback and question-answering systems~\cite{IDEAL_14}. A comprehensive review of ontology use in e-learning systems can be found in~\cite{Ontologies_Al_Yahya} . 

In the educational area, in terms of technical solutions to facilitate ontology building, authoring tools for ontology creation dominate the research field (e.g.~\cite{Yang2004},~\cite{Aroyo2004}), while semi-automatic~\cite{Zouaq2009} and automatic~\cite{Reasoning_Henze} approaches are less researched. In the wider ontology development field, there are tools for semi-automatic (e.g.~\cite{Kamel2013}) and automatic (e.g.~\cite{Text2Onto}) ontology building, however, these tools were designed for IT experts, not educators~\cite{Ontology_Extraction_hatala}. 

Recent research in learning technologies took up existing semantic web knowledge and applied it to improve learning environments. This research includes educational data mining based on semantic web~\cite{enterprise_nayak}, integrating educational resources with service-oriented architectures and web services using semantic web~\cite{Measuring_Jianhua}, and semantic web applications for education~\cite{education_kasimati}. 

In this paper we propose an approach to automatically build a general subject ontology (i.e. a domain ontology for an academic subject), for educational purposes, from textual resources, by leveraging data mining techniques. Unlike previous research, both in the educational domain and the wider ontology building area, we use overlapping textual resources to overcome the \textit{tf-idf} approach limitations. Moreover, while most of the previous research used linguistic approaches that require manually built term lists or pre-defined lexico-syntactic patterns, we propose a frequent pattern mining approach that does not require these, which makes the proposed approach domain-independent and generates a connected acyclic concept graph. To the best of our knowledge, this is the first use of a frequent pattern mining approach to ontology building.   

The resulting ontology serves as a knowledge source for a question-answering system. To the best of our knowledge, there are no question-answering systems for education underpinned by automatically generated ontologies. The proposed question-answering system was validated by domain experts using convenience sampling~\cite{Research_Frederick}. Also, we used different semantic similarity metrics to validate the returned answers and identify a suitable metric for wider validation (without the need for information from experts). Our experiments show that the Latent Semantic Analysis (LSA)-based text similarity metric is the most suitable metric for validating the question-answering results. 

We validated the subject ontology learning system through the results of the questions-answering system. We used a comparative validation approach by comparing the results when using our ontology with the results when using an ontology generated by Text2Onto~\cite{Text2Onto}\footnote{Text2Onto standalone version released on 09/11/2007, available at http://ontoware.org/projects/text2onto/}, one of the most popular tools for ontology learning from textual resources. This measure of usefulness of the ontology opens the door to new objective ways of assessing %the goodness of 
a given ontology, when compared with the long practice of subjective assessment by domain experts.

\begin{figure}[!t]
\centering
\begin{tikzpicture}[node distance=1pt,
blueb/.style={
  draw=white,
  fill=mybluei,
  rounded corners,
  text width=2.5cm,
  font={\sffamily\bfseries\color{white}},
  align=center,
  text height=8pt,
  text depth=6pt},
greenb/.style={blueb,fill=mygreen},
]
\node[blueb, text width =2cm] (RCP) {Rules};
\node[blueb,below=of RCP, text width= 3cm] (RTe) {Relations};
\node[blueb,below=of RTe, text width=4cm] (Jti) {Concept Hierarchy};
\node[greenb,below= of Jti,text width=6cm] (las) {Concepts};
\node[greenb,below= of las,text width=7cm] (syn) {Synonyms};
\node[greenb,below= of syn,text width=8cm] (term) {Terms};
\node[font=\sffamily\itshape\color{white},above=of RCP]{};
\pgfdeclarelayer{background}
\pgfdeclarelayer{foreground}
\pgfsetlayers{background,main,foreground}
\begin {pgfonlayer}{background}
\draw[blueb,draw=black,fill=mybluei!30] 
  ([xshift=-2pt,yshift=2pt]current bounding box.north west) rectangle 
  ([xshift=2pt,yshift=-2pt]current bounding box.south east);
\end{pgfonlayer}
%\node[blueb,draw=black,fill=myblueii,below=4.8cm of Bro,text width=13cm+44pt] (RCP) {RCP Runtime};
\end{tikzpicture}
\caption{Ontology learning layer cake}
\label{OntoCacke}
%\vspace{-15pt}
\end{figure}
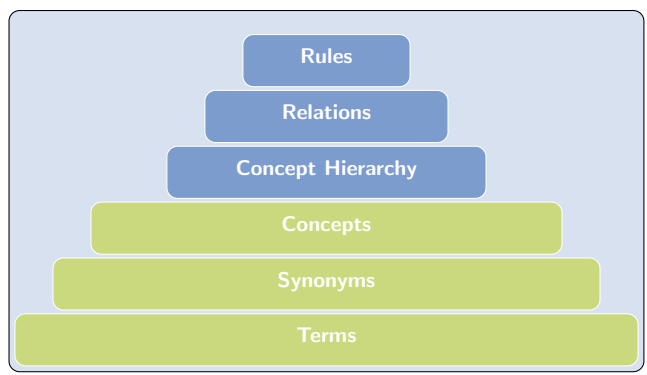

The rest of the paper is organised as in the following. Section 2 presents the related work to our research and Section~3 describes our proposed ontology developing system, including a description of the approach and the implementation results. The question-answering system is presented in Section 4, while Section 5 describes %our experimental results and analysis.
the validation of our question-answering system underpinned by the automatically built ontology. % is described in Section 6,
Section 7 concludes the paper and discusses directions for further research.

%****************************************************************************************************************************************************************************************************************************************************************************************************************************
\section{Background}
The term ``ontology'' has been used in the knowledge engineering community to describe ``formal explicit specifications of shared conceptualisations'' \cite{Atranslation_Gruber}. Ontologies are used to formally model a domain and represent complex information into machine-readable format.
%Ontologies formally model an underlying domain structure using a machine readable format. Also, they require a consensual view of the underlying domain. 

Ontologies play a key role in the Semantic Web, which was introduced by Tim Berners-Lee \cite{Berners-Lee2001} to establish common terminologies among software agents. The aim was to enable different software agents to have a shared understanding of terminology. 
%\deleted{The process of building an} 
Ontology development is a knowledge engineering task, which requires extensive effort and time. Researchers proposed a number of methods for ontology development in the quest to reduce the complexity of the ontology building process. As a result, the ontology learning field emerged.

Ontology learning is concerned with knowledge acquisition. It consists of several phases which are: terms extraction, finding synonyms, concepts identification, concept hierarchy construction, relations discovery, and sets of rules derivation~\cite{Aroyo2006}. Fig.~\ref{OntoCacke} shows the general ontology learning layer cake~\cite{Buitelaar2005} outlining the phases mentioned above. 

%\deleted{
Subject ontologies aim to make a subject knowledge explicit. Most of the relations among subject-knowledge concepts are in the form of ``is--a'' relationship or its inverse ``has subtype'' relationship \cite{Boyce_Education}. Also, the axioms can be moved to applications (agents) that utilise the subject knowledge.
As a result, the ontology layer cake can be reduced to the four bottom layers. 
Working with just these layers would facilitate a general ontology learning process, while also allowing the explicit representation of subject knowledge. 
%}

A number of ontology learning researchers explored Natural Language Processing (NLP) techniques to discover domain concepts and relationships among concepts from unstructured text documents \cite{Anapproach_Valencia}, \cite{NLP_Maynard}, \cite{Velardi2005}, \cite{OntoGain}. The Text2Onto \cite{Text2Onto}, OntoGain \cite{OntoGain}, and OntoLearn\cite{Velardi2005} systems used NLP tools to extract key concepts from text. They used the \textit{tf-idf} measure  for the selection of domain ontology concepts; this metric, however, is sensitive to the size and specificity of the documents used, thus having limitations in relation to the identification of domain-specific concepts~\cite{Wong2007,Balachandran2016}. 
To overcome these limitations, we use overlapping text resources, which include but are not limited to, short documents containing intensive domain-specific concepts i.e. slide notes and instructor's notes, thus addressing the size (by having multiple documents) and specificity (by using documents rich in domain information). To further distinguish domain from non-domain knowledge we filter terms that are frequent in the corpus of contemporary American English (COCA)~\cite{COCA}, i.e. commonly used terms (non-domain specific).

%\begin{table}[!t]
%\caption{Ontology learning}
%\label{prevWork}
%\centering
%\begin{tabular}{|l|l|l|l|l|}
%\hline
%Reference & Automatic & Unstructured text & Lexical/syntactic knowledge & Techniques \\
%
%\hline
%\end{tabular}
%\end{table}

A number of previous works have employed association rules algorithms for identification of frequent patterns. The predictive Apriori algorithm in combination with a probabilistic algorithm was used in~\cite{OntoGain} to identify non-taxonomic relations, i.e. non-hierarchical relations.
Similarly, a variant of the generalized association rule mining algorithm was used for mapping non-taxonomic relations~\cite{jiang_crctol}. Unlike these works, we use frequent pattern mining (the FP-Tree algorithm) for the identification of hierarchical/taxonomic relations.

%lexico-syntactic methods

A statistical modelling method for structured prediction called Conditional random fields (CRF) was used in~\cite{espinosa_extasem} to identify hierarchical relations. This approach requires manual annotation to train the CRF model, which can then be used for identification of relations. Unlike this approach, we use an unsupervised algorithm which does not require manual annotation for the identification of hierarchical relations.

Many other ontology learning approaches have been proposed from unstructured text, e.g.~\cite{Zouaq2011,Arabshian2012,Adding_chen,Mining_Viral}; unlike these, our approach does not require any pre-defined domain specific term lists or lexico-syntactic rules.

%Unlike the aforementioned systems, the proposed approach does not require any pre-defined domain specific term list or lexico-syntactic rules. Also, researchers used semantic similarity to support the ontology learning process. For example, \cite{Adding_chen} used Latent Semantic Analysis (LSA) to support the concept discovery. \cite{OntoGain} used probabilistic and rule based techniques to discover non-taxonomic relations and they used hierarchical text clustering and formal concept analysis for extracting taxonomic relations. While recent research utilised the structure of web pages~\cite{Effective_ahmed} or the formatting structure of learning documents and book outlines and indexes \cite{automatic_larranaga} to discover the underlying concepts and properties of a domain. They leveraged the Wikipedia/subject document structure to retrieve concept definitions, and identify existing relationships. Mining domain specific glossaries and texts to enrich and evaluate ontologies has also been proposed~\cite{Mining_Viral}. Extending a pre-defined domain terms by leveraging large semantic networks for constructing a domain taxonomy only made up of semantically pertinent edges was carried by \cite{espinosa_extasem}.

Ontologies have been used in the educational field to represent course content~\cite{An_Intelleginet_Rebecca,Boyce_Education,Zouaq2009}. It can scaffold students learning due to its role in instructional design and curriculum content sequencing \cite{supporting_cesar}. Also, ontologies have been used in intelligent tutoring systems \cite{An_Intelleginet_Rebecca}, student assessments \cite{Kate_Ontology}, and feedback \cite{Pedro_An_ontology,IDEAL_14}. An ontology-based feedback framework to support students in programming tasks was introduced by \cite{Pedro_An_ontology}. They suggested a framework for adaptive feedback to assist students in programming. The framework aimed to help students in correcting programming syntax errors. In spite of describing their work as ontology-based feedback, they did not describe the structure of their ontology nor the process of creating that ontology (manual/automated). 

%\deleted{The educational field employs ontologies to enhance teaching and learning in both traditional and e-learning settings. However, developing ontologies for educational purposes is one of the limitations in adopting educational ontology-based applications since developing an ontology is notoriously costly and time-consuming. 
%To tackle ontology developing limitations, many tools have emerged to automatically or semi-automatically extract and build ontologies from text \cite{Text2Onto}, \cite{O4E}, \cite{Zouaq2009}. There are also a number of tools that leverage course content resources to automatically or semi-automatically build course subject ontologies \cite{Text2Onto}, \cite{O4E}, \cite{TM4L_Dicheva}.
%According to a relatively recent paper \cite{Ontology_Extraction_hatala}, existing ontology tools are categorised into three categories, which are: hand-crafting ontologies from scratch, semi-automatic ontology building, and search and retrieval of ontology from online resources. The authors asked educators to evaluate the use of tools for semi-automatic or automatic ontology building, and their results indicated that the ``current state of the tools for developing domain ontologies by educators is unsatisfactory''(\cite{Ontology_Extraction_hatala}, p. 13).}

To validate an ontology, a structural or a functional approach can be used~\cite{Velardi2005}, \cite{abstract_verma}. %\added{
The former typically involves a comparison of the resulting ontology against a pre-defined (gold-standard) ontology; however, there is no approved way of evaluating ontologies based on gold-standard evaluation~\cite{Dellschaft_onHow}. %}. 
The latter involves the use of the ontology for a particular task and measuring the impact of the ontology use on this task.   
We employ the functional approach using the resulting ontology in a question-answering system for MOOCs, a functionality highly required in such courses with thousands of enrolled students. 

A typical question-answering system aims to automatically answer user questions which are asked in a natural language syntax. Educational question-answering systems are limited  due to the poor quality of the returned answers~\cite{AnIntelligent_Feng}. On the other hand, general question answering systems return good quality general answers~\cite{start_katz}. Domain specific questions typically result in inaccurate answers due to the limitations of NLP approaches based on linguistic information~\cite{Gupta2008,Question_Moll}. 

A notable exception is the approach in~\cite{Poon2010}, which used an unsupervised approach for ontology building employing hierarchical clustering and then using the ontology to answer questions related to the medical domain. They constructed the ontology using the GENIA dataset (containing PuBMed abstracts) and created simple questions by sampling verbs and entities according to their frequencies in GENIA~\cite{Poon2010}. They obtained a very high accuracy of 91\% for the question-answering; however, this may be due to the use of simple questions, which the authors argued were chosen to focus the evaluation on the knowledge extraction, rather than the question handling. 

The techniques mentioned above are not efficient for real-time learning environments, especially MOOCs, due to the large volumes of questions involved. Recently, question-answering systems for education and especially for online learning environments have emerged~\cite{Instructor_Dunwei,IDEAL_14}. With the exception of our previous work~\cite{IDEAL_14}, ontologies were not used in this research area. 

% \added{
We propose an unsupervised ontology learning framework for unstructured educational text document collections. We aim to make knowledge, existing in these documents, explicit, which in turn, enables natural language applications such as question answering systems. We tackle the ontology learning problem as a data mining task by leveraging educational document characteristics such as cohesion, isolation, and unity.
%}

%\deleted{We propose a generic automatic ontology development system for subject learning contents.} Unlike the work reported in \cite{Zouaq2009}, \cite{Text2Onto}, it extracts key terms from an overlapping collection of subject learning content resources. % using term relative frequency weights. 

We use a regular expression parser widely adopted in programming language compilers in a new paradigm to index a course learning content and to identify the learning content concepts instead of using linguistic analysis techniques or predefined lexico-syntactic patterns typically used, e.g. \cite{Zouaq2009}, \cite{OntoGain}, \cite{espinosa_extasem}, \cite{velardi_ontolearn}, \cite{HayesCollaborative}. 

Finally, we propose a novel approach to build the concept hierarchy for the course learning contents. Unlike the aforementioned approaches in the literature~\cite{Text2Onto}, \cite{Zouaq2009}, \cite{TM4L_Dicheva}, \cite{HayesCollaborative}, \cite{taxofinder_kang}, we use the frequent patterns and the term association techniques to build the concept hierarchy for learning contents by customising the FP (Frequent Pattern)-Tree structure~\cite{Han_MiningFP}. We use a heuristic function to enhance the quality of the generated concept hierarchy by ensuring that each concept appears in the hierarchy only once; term associations drive this heuristic function. To the best of our knowledge, no previous work uses the FP-tree algorithm for creating the concept hierarchy, nor the term associations to automatically solve the issue of a concept appearing multiple times in the hierarchy. 

The techniques mentioned above allow the proposed system to work across different subjects since it does not require any specific lexico-syntactic information. The next section explains the ontology building process in details.

%**************************************************************************************************************************************************************************************************************************************************************************************************************************************************************
%\vspace{-5pt}
\section {Phase I: Ontology Building}

In this section we present our proposed approach to automatically develop a subject ontology. %\deleted{We start by formally defining the general domain ontology, then} 
First, we present our definition for a subject ontology and the purpose of developing the subject ontology. The proposed approach is described in detail for all the stages involved in the process.

An ontology is an explicit formal specification of a shared conceptualisation of a domain of interest \cite{staab_Handbook_on_ontology}. An ontology defines the intentional part of the underlying domain, while the extensional parts of the domain (knowledge itself or instances) are called the ontology population. %\added{
Ontologies are categorised as formal ontologies, prototype-based ontologies, and terminological ontologies \cite{sowa_knowledge}. In this research, we build a terminological ontology which it construct subtype-supertype relations and describe concepts by labels or synonyms.%}

%Definition~\ref{onto1} formally defines an ontology~\cite{hotho_Ontology}.

%\begin{onto}
%\label{onto1}
%A core ontology is a sign system $ \Theta :=(T, P, C^*, H, Root)$, where\\
%$T$: a set  of  natural  language  terms  of  the  Ontology\\
%$P$: a set of properties\\
%$C^*$: a function that connects terms t $\in T$ to a set $p \subset P$ \\
%$H$: a hierarchical organisation connecting all terms from $T$ in acyclic, transitive, directed relationships.\\
%$Root$: is the top level node where all concepts in $ C^* $ are mapped to it.\\
%\end{onto}

Different ontology learning systems and methods address different ontology learning tasks. As a result, it is difficult to compare these methods. In consequence, ontology learning for education presented different perspectives and had different purposes. So, to clarify our methodology and to build a common background for this research we will define our proposed ontology, identify its purpose, and introduce our motivation for developing a subject ontology.

\emph {Definition}:
A subject ontology is a formal representation of the contents of a particular academic subject that makes knowledge explicit.

\emph {Purpose}: 
Learners consume learning contents to get knowledge. We aim to formally represent the contents of a particular subject to scaffold technology enhanced learning systems in delivering course contents to learners. In particular, we aim to answer content-related questions.

\emph{Motivation}:
The massiveness property of MOOCs makes it difficult for the course facilitators to answer learners' questions in a timely manner. This increases the learners' cognitive load and may increase the drop-out ratio. This motivated us to develop a general subject ontology to underpin an automatic answering system for the learners' content-related questions.

\begin{figure}[t]
\centering
\resizebox{\textwidth}{!}{
\includegraphics{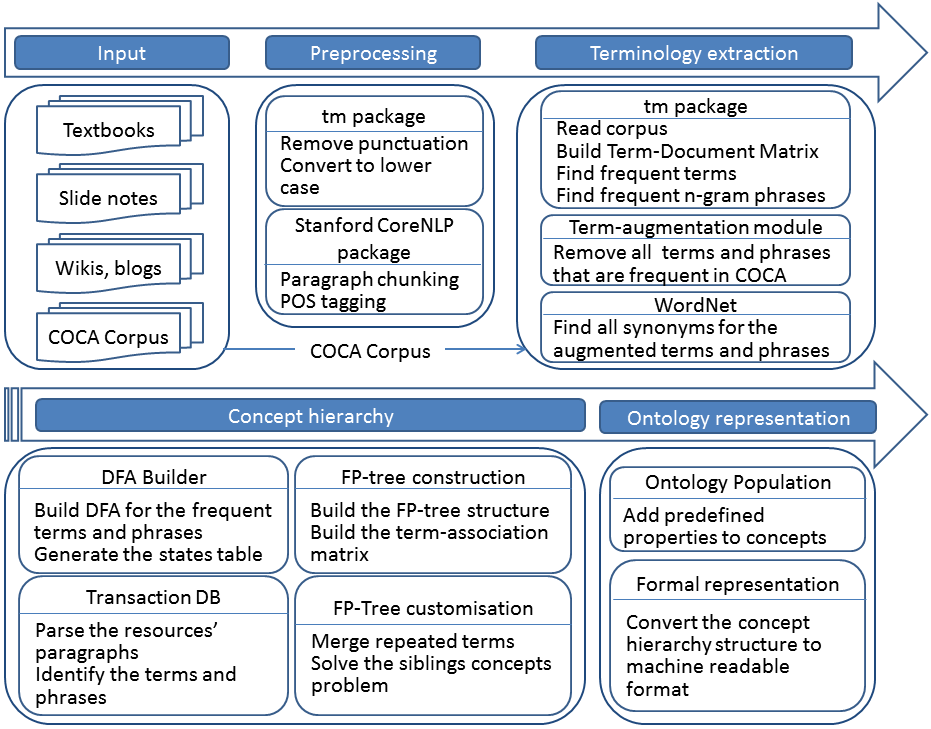}
}
%\vspace{-20pt}
\begin{minipage}{\textwidth - 20pt}
    \footnotesize
    %\emph{YOUR NOTES}
		COCA = the corpus of contemporary American English;\\
		POS = part of speech;\\
		DFA = deterministic finite automata;\\
		FP-tree = frequent pattern tree;\\
		DB = database.
\end{minipage}
\caption{The proposed ontology development system}
\label{sa}
%\vspace{-15pt}
\end{figure}

Fig.~\ref{sa} illustrates the proposed ontology development system. It shows the different phases to build a subject ontology and the packages we used or developed in every phase: (1) identify subject resources; (2) preprocess the data resources; (3) extract the subject terms; (4) construct the concept hierarchy and apply our proposed heuristic function to enhance the quality of the concept hierarchy; and (5) export the concept hierarchy into a formal representation. In the following subsections we will describe these phases in details. 

%*****************************************************************************************************************************************************************************************************************************************
%\vspace{-5pt}
\subsection{Data Collection and Preprocessing}

In the educational domain, ontology learning research typically uses textbooks' table of contents, the structure of web pages or text formatting hierarchies to extract the underlying subject terms and to build the concept hierarchy for the underlying domain~\cite{TM4L_Dicheva,Zouaq2009,Boyce_Education}. Many online and traditional educational resources, however, lack any given structure or text formatting hierarchy. As a result, the existing tools and techniques are not appropriate for these resources. We address this issue by building %our assumption to build
on the assumption that a subject ontology can be derived from heterogeneous overlapping learning objects (LOs) resources. These resources include textbooks, lecture notes, blogs, and other plain text subject resources. In this context, we do not need any knowledge about terms and the relationships among these terms, thus overcoming the limitation of lack of structure. This also allows a general approach to ontology building, from which ontologies for a variety of subjects can be built.

Generally, in the didactic domain, educators share a set of specific concepts for a subject's knowledge. As a result, when we collect overlapping resources for a subject, we can reveal that subject's concepts. 

Educational documents provide definitions and explanations about concepts to be learnt. These concepts have typically low ambiguity and high specificity -- for this reason learning objects are good candidates for building a subject course ontology. Textbooks share some characteristics when grouping concepts together in learning units \cite{Rakesh_Toward}. These characteristics support the proposed approach of learning ontologies. These characteristics are:

\begin{itemize}[leftmargin=*]

\item{Cohesion:} Each learning unit consists of concepts that are closely related. For example concepts like ``data'', ``information'', and ``knowledge'' are closely related and appear together in the ``Introduction to Database'' course for instance. While, ``Normalisation'',``Concurrency Control'', and ``DML'' are not tightly connected. As a result, related concepts appear together in learning units.
\item{Isolation:} Concepts that belong to different learning units must be independent as much as possible.

\item{Unity:} Some concepts, especially fundamental ones, may appear in different learning units.
\end{itemize}

In order to test our proposed system, we collected overlapping learning objects for the ``Database Design and Management'' subject.
% \footnote{Database Systems: Applicational Approach to Design, Implementation, and Management; 4th Edition.}$^{,}$ \footnote{Database Management Systems; 2nd Edition.}$^{,}$\footnote{Fundamentals of Database Systems;6th Edition}. 
These resources are a combination of book chapters~\cite{Ramakrishnan2000,Elmasri2010}, slide notes, blogs, and Wikis\footnote{https://en.wikipedia.org/wiki/Database}. All resources are stored in plain text format. 

%*****************************************************************************************************************************************************************************************************************************************

\subsection{Term Extraction}
Terminology extraction is the process of discovering terms that are good candidates to represent the underlying domain in an ontology. It is the first and an important step in developing a domain ontology. Arguably, this is a matured phase and a plethora of techniques and measures exist in the literature. However, term extraction for education ontologies has not been examined to determine the best technique for developing a subject ontology. Thus, we use one of the most popular approaches based on metrics of term frequency (\textit{TF}), as well as $n$-grams to ensure the extraction of more complex terms. In accordance with the W3C standard\footnote{https://www.w3.org/standards/semanticweb/ontology}, we use terms and concepts interchangeably.  
%In this research we used the term frequencies \textit{(TF)} and $n$-gram techniques to extract the key terms. %\added{
%\textit {TF} measure is simple but effective and outperforms other statistical measures.
 
The relevance of a term to a domain depends on the \textit{TF} measure, but also, the performance of the concept extraction methods is highly affected by the size and specificity of the documents collection used for ontology learning. As pointed out in the introduction, approaches based on \textit{tf-idf} are not effective on small numbers of documents, and have limitations in terms of identifying domain-specific concepts~\cite{Wong2007,Balachandran2016}. 
Given the aforementioned characteristics of educational resources, we proposed two approaches in order to overcome these problems: (a) using overlapping text resources and (b) using resources rich in domain-specific information. We used the \emph{``tm''} and \emph{``RWeka''} packages in R\footnote{https://www.r-project.org/} to process the subject learning resources \cite{tm,RWeka}. Also, we used the COCA corpus (collection of documents) to filter out the frequent common-language (i.e. non domain-specific) terms~\cite{COCA}.

%*******************************************************************************************************************************************
\subsubsection{Approach}
First, we built the document-term matrix (DTM) which is a two dimensional array data structure. A DTM describes the frequency of terms that occur in a corpus. Usually, rows correspond to words in the corpus and columns correspond to documents in the corpus. The cell value describes the frequency of a word in a given document. 

%\added{
Given an extracted term \textit{t} in a document set \textit{d}, %}
 we used term frequency-inverse document frequency (\textit{tf-idf}) as is described in Equation \ref{tf_ifd}, indicating the importance of a word in a document~\cite{Chowdhury2010}, as the frequency weighting scheme. 

\begin{equation}
\label {tf_ifd}
\text{tf-idf}_{t,d} \; = \; tf_{t,d} \times idf_t \; = \; tf_{t,d}
        \times log(\frac{N}{df_t})
\end{equation}
where \textit{tf} stands for term  frequency, $df_t$ is the count of documents where t appears, $idf$ stands for inverse document frequency, and $N$ is the number of documents in a corpus.

All terms with frequencies above a given threshold $\theta$ are extracted as potential candidates for subject terms. Experimentally, we found that a value of 0.90 for  $\theta$ gives the best list of subject terms. Since we use overlapping subject resources, we expect to identify most of the subject key terms this way. This approach is appropriate for educational documents, and our experimental results reported in Section~5 support this claim.

When we used the frequency measure to identify the ontology terms, we retrieved many irrelevant terms. In order to overcome this drawback, we augmented the obtained terms based on the following approach: we assumed a term is a good candidate for a domain ontology if the term's \textit{tf-idf} value is greater than the term's \textit{tf-idf} in the corpus of the daily used terms. To achieve this, we used the Corpus of Contemporary American English (COCA) to get all frequent daily-used terms and phrases. The COCA corpus has more than 189,431 texts in the 450+ million word corpus (the last update to the corpus was in June 2012) \cite{COCA}. As a result, any frequent term in the underlying subject corpus that is not one of the frequent terms in the COCA corpus is a candidate term for the subject ontology. 

We repeated the same approach with frequent $n$-gram terms, where $n$ is the number of the words in a term ($2 \leq n \leq 5$). We set the maximum $n$-gram phrase to 5 words since our experiments showed that $n$-gram phrases that have 6 or more words are not frequent in the corpus even when we reduce the threshold $\theta$ to lower values. We extracted the $n$-gram phrases using \emph{``RWeka''} package in R \cite{RWeka}. 
Every frequent $n$-gram term in the underlying subject corpus which is at the same time not a frequent $n$-gram term in the COCA corpus becomes a candidate term for the subject ontology.

In the next step, we used  ``Jawbone Java API'' through the Wordnet package in R to identify all synonyms for the candidate terms. Wordnet is a large English lexical database. It groups nouns, verbs, adjectives and adverbs into sets of cognitive synonyms called synsets, where each synset expresses a distinct concept. Synsets are interlinked by means of conceptual-semantic and lexical relations~\cite{WordnetJ}, \cite{WordnetR}.

A possible disadvantage of this approach, like for many automated approaches, is that some concepts which are related to the subject ontology may not appear in the extracted terms. %However, we can allow educators or even learners to add any missing terms which is a task that does not require any technical expertise and can be achieved through a simple user interface.

Algorithm \ref{A1} shows the pseudo-code for retrieving the subject ontology terms. The algorithm takes the subject course as input and returns a list of candidate terms and their synonyms.
\begin{algorithm}
\caption {Extracting subject ontology terms}
\label {A1}
\begin{algorithmic}[1]
\Procedure{FrequentTerms}{corpus, terms}
\State $terms \gets \emph {null}$
\State $\textit{$\Theta$} \gets threshold$
\State $\textit{ COCA} \gets {\textit{Corpus of Contemprory English}}$
\State $\textit{DTM} \gets {\textit{document terms matrix(corpus)}}$
\State $\textit{terms} \gets {\textit{freq terms(DTM,Tf-Idf,$\Theta$) }}$
\For {\texttt{{($k = 2$,$k < 6$, $k{+}{+}$)}}}
	\State $\texttt {terms} \gets {terms \bigcup freq(n-gram(DTM,k),\Theta)}$
\EndFor
\For {\texttt{($t \in terms$)}}

	\If {$\textit {freq (t) $\textless$ COCA(t)}$}
		\State $\texttt terms \gets{terms -t}$
	\EndIf
\EndFor
\State $\textit{terms} \gets wordnetSynonyms(terms)$
\EndProcedure
\end{algorithmic}
\end{algorithm}

%*********************************************************************************************************************************************
\subsubsection{Implementation}\label{termExtractionImplem}
%*****************************************************************************************************************************************************************************************************************************************
%\vspace{-5pt}
%\subsection{Term Extraction}
The system found all frequent words in the corpus, as well as bigram, trigram, 4-gram, and 5-gram frequent phrases. All frequent terms that are not frequent in the COCA dictionary were selected to represent the subject ontology as described in Algorithm \ref{A1}.

The ``Wordnet'' library was used to retrieve all possible synonyms of the extracted concepts. We found that this step generated many irrelevant terms. A possible reason is that terms and concepts in a subject domain are used in more specific contexts than their general meaning. For example, the term ``table'' is used to describe the data structure for storing data in relational databases; however, synonyms like ``bench'', ``worktop'' or ``counter'' are not used in the context of the relational database subject to describe the same data structure. 

These extra synonyms did not significantly affect the quality of the domain ontology, but resulted in an increase of computation complexity of the subsequent steps. Table~\ref{t2} shows a subset of the terms extracted after implementing this phase.

\begin{table}[!h]
\caption{Sample of extracted concepts for the ``Database Design and Management'' subject}
\label{t2}
\centering
\begin{tabular}{|c|l|}
\hline
ID & Term  \\ \hline 1 & root  \\ \hline 2 & data  \\ \hline 3 & data file  \\ \hline 4 & data independence  \\ \hline 5 & data item  \\ \hline 6 & data model  \\ \hline 7 & data types  \\ \hline 8 & data warehouse  \\ \hline 9 & database  \\ \hline 10 & database application  \\ \hline 11 & database management  \\ \hline 
\end{tabular}
\end{table}

%
%\begin{table}[!t]
%\caption{Subset of the Database Design and Management terms}
%\label{t1}
%\centering
%\begin{tabular}{|c|l|}
%\hline
%ID & Term  \\ \hline 1 & backup  \\  \hline 2 & calculus  \\ \hline 3 & client server  \\  \hline 4 & commit  \\ \hline 5 & conceptual data  \\ \hline 6 & conceptual schema  \\ \hline 7 & concurrency  \\ \hline 8 & concurrency control  \\ \hline 9 & data  \\ \hline 10 & data entry  \\  \hline 11 & data model  \\ \hline 12 & data structures  \\ \hline 13 & data types  \\ \hline 14 & data warehouse  \\ \hline 15 & database  \\ \hline 
%21 & database application  \\ \hline 22 & database design  \\ \hline 23 & database management  \\ \hline 24 & database schema  \\ \hline 25 & dba  \\ \hline %26 & dbms  \\ \hline 27 & ddl  \\ \hline \hline 13 & data file  \\ \hline 14 & data independence  \\ \hline 15 & data item  \\
%\end{tabular}
%\end{table}

%*****************************************************************************************************************************************************************************************************************************************

\subsection{Concept Hierarchy Construction (Taxonomy learning)}

Taxonomy learning is the process of building a taxonomy by identifying the underlying domain-specific concepts and their taxonomic relations from the domain text corpus. Taxonomies play an important role in developing successful application for the underlying domain \cite{taxofinder_kang,semantic_Meijer}.
 In the ontology learning field, a number of research projects used syntactic and semantic techniques to extract hierarchical relationships among the concepts of the underlying domain \cite{Text2Onto}, \cite{Anapproach_Valencia}, \cite{taxofinder_kang}. However, recently there is a growing trend toward using machine learning techniques to determine relationships among concepts. For example, for ontology learning, association rules have been used for identifying~\cite{OntoGain} and filtering~\cite{jiang_crctol} non-hierarchical relations; conditional random fields (CRF) were also used~\cite{espinosa_extasem} to identify hierarchical relations.

%Researchers used Support Vector Machines \cite{Using_Li}, Maximum Entropy Models \cite{Research_Meng} and Hidden Markov Models \cite{Information_Freitag}, modified generalised association rule mining \cite{jiang_crctol} to name a few.

One of the most popular algorithms is Apriori, which identifies frequent subsets of items (i.e. itemsets) which are common in transactional databases. This algorithm requires the generation of candidate itemsets (a computation-intensive process) and the calculation of a measure of their frequency (i.e. support) which is then used to filter out (or prune) infrequent itemsets. Unlike Apriori, the FP-Tree algorithm does not require the time-consuming generation of candidate itemsets.

In this research, we used data mining techniques to extract the hierarchical relationships among concepts. We leveraged the characteristics of a subject course resources where intuitively related topics are grouped together or appear together in the learning resources (cohesion property). Specifically, we customised the frequent-pattern tree (FP-Tree) structure which was proposed by Han et al. and defined as in Definition~\ref{d2}~\cite{Han_MiningFP}. 
\begin{onto}
\label{d2}
A Frequent Pattern Tree (FP-Tree) is a tree structure defined as follows:
\begin{description}
    \item[A]{It has one root node, a set of item-prefix subtrees as the children of the root, and a frequent-concept header table.}
    \item[B]{Each node in the item-prefix subtrees consists of three fields:}
\begin{enumerate}
    \item{item name: registers which item is represented by the node;}
    \item {occurrence frequency: the number of transactions represented by the portion of the path reaching the node; and} 
    \item {node-link: refers to the next node in the FP-tree carrying the same item, or null if there is none.}
\end{enumerate}
    \item[C]{Each entry in the frequent-concept header table consists of two fields: (a) item name and (b) head of node-link, which points to the first node in the FP-tree carrying the item.}
\end{description}
\end{onto}
An FP-tree is a compact structure that stores quantitative information about frequent patterns (see Definition~\ref{d1}), i.e. frequent sets of items (called itemsets), in a transaction database; it stores items and their frequencies. In our application, the items are concepts.

In order to build an FP-Tree, we need a transaction database (DB) and a minimum support threshold $\theta$, as defined in Definition~\ref{d1}. We considered every paragraph in the corpus as a transaction. All distinct concepts that appear in a paragraph form the transaction items. To capture all concepts we set $\theta$ to Zero.
\begin{onto}
\label{d1}
 Let C=\{$c_1, c_2,\dots,c_m$\} be a set of concepts of a particular course.\\
DB=\{$T_1,T_2,\dots,T_n$\} a Transaction Database, where $T_i$ ($i\in[1..n]$) is a transaction that contains a set of concepts $\in$ C.\\
Let Support (S) be an occurrence frequency.\\
Let $\theta$ be the minimum support threshold.\\
Then, P is a frequent pattern $\implies$ (P is a set of concepts $\in$ C) $\land$ S(P) $\textgreater \theta$.
\end{onto}

In order to generate the transaction database for the subject course, we split the corpus into a set of paragraphs using the \emph{``openNLP''} package for R \cite{openNLP}. The \emph{``openNLP''} library is a machine learning based toolkit for processing of natural language texts written in Java. It supports the most common NLP tasks, such as tokenisation, sentence segmentation, part-of-speech tagging, named entity extraction, chunking, and parsing. Also, we used the Stanford coreNLP library for co-reference resolution \cite{CoreNLP}. We parsed each paragraph in the corpus, and, as a result, we extracted the concepts appearing in that paragraph through the procedure explained in the following subsection.\\

%\noindent \textit{\textbf{DFA Builder\\}}\label{dfa}
%\paragraph{DFA Builder}
%************************************************************************************************************************************************
\subsubsection{DFA Builder Approach}\label{dfa}

In order to extract the concepts that appear in a paragraph, we parse the paragraph word by word to discover all terms in a paragraph. To parse a paragraph, we built a deterministic finite automata for every term or concept extracted from the subject course textual resources. We considered every concept or any possible synonym a deterministic finite automata (DFA). 

DFA is formally defined in Definition \ref{d21}. In our approach, $\Sigma$ is the set of all natural language words which are selected to represent a subject ontology. We developed an automated DFA generator module that takes all concepts and their synonyms as input and generates a DFA for every concept and its synonyms. 

\begin{onto}
\label{d21}
A deterministic finite automaton (DFA) is a 5-tuple: (Q, $\Sigma$, $\delta$, q0, F), where Q is a finite set called the states, $\Sigma$ is a finite set called the alphabet, $\delta$: Q $\times \Sigma$  $\rightarrow$  Q is the transition function, q0 $\in$ Q is the start state, and F $\subset$ Q is the set of accept states.
\end{onto}

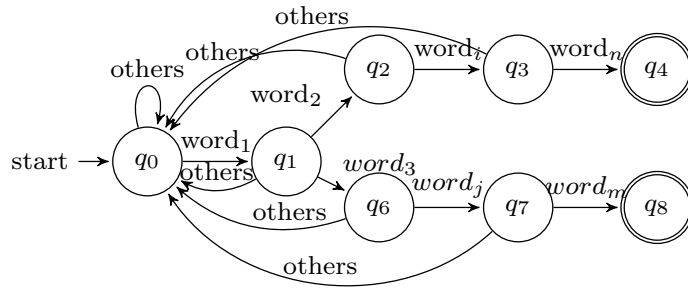
\begin{figure}[!t]
\centering
\resizebox{\textwidth}{!}%
{
\begin{tabular}{|l|}
\hline \\
\subfloat[Deterministric Finite Automata for concept $c_1$] 
{
\resizebox{\textwidth}{!}%
{
\begin{tikzpicture}[>=stealth',shorten >=1pt,auto,node distance=1.5cm,baseline]
	\node [align =left] at(-1,0) (bb)   {Let \emph{c1} be a concept in a subject course ontology. \\ Let \emph{n} be the number of words in \emph{c1}. \\
Let \textgravedbl others\textacutedbl  be any word $\notin$ \emph{c1} words \\Then the DFA that represents \emph{c1} is:\\};
  \node[initial,state](q0)  at(-3,-2)  {$q_0$};
  \node[state]         (q1) [right of=q0]  {$q_1$};
  \node[state]         (q2) [right of=q1] {$q_2$};
  \node[state]         (q3) [right of=q2] {$q_3$};
  \node[state,accepting]         (q4) [right of=q3] {$q_4$};
  \path[->] (q0)  edge [loop above] node {others} (q0)
             edge              node {word$_1$} (q1)
        (q1) edge [bend left]  node {others} (q0)
             edge              node {word$_2$} (q2)
        (q2) edge [bend left]  node {others} (q0)
			    edge              node {word$_i$} (q3)
        (q3) edge [bend left]  node {others} (q0)
			    edge              node {word$_n$} (q4)
        ;
	\node [align =left] [below of =q4] (bbb)  {\\where $q_4$ is a final state.};
\end{tikzpicture}
}}
%\hline 
\\
%\par
\subfloat[Deterministric Finite Automata for the Concept $c_2$] 
{
\resizebox{\textwidth}{!}%
{
\begin{tikzpicture}[>=stealth',shorten >=1pt,auto,node distance=1.5cm,baseline]
	\node [align =left] at(-1,0) (bb) {Let \emph{c2} be a concept in a subject course ontology. \\ Let \emph{m} be the number of words in \emph{c2}. \\
Let others any word $\notin$ \emph{c2} words \\Then the DFA that represents \emph{c2} is:};
  \node[initial,state](q0) at(-3,-2)   {$q_0$};
  \node[state]         (q5) [right of=q0]  {$q_5$};
  \node[state]         (q6) [right of=q5] {$q_6$};
  \node[state]         (q7) [right of=q6] {$q_7$};
%  \node[state]         (q3) [right of=q2] {$q_7$};
  \node[state,accepting]         (q8) [right of=q7] {$q_8$};
  \path[->] (q0)  edge [loop above] node {others} (q0)
             edge              node {word$_1$} (q5)
        (q5) edge [bend left]  node {others} (q0)
             edge              node {word$_3$} (q6)
        (q6) edge [bend left]  node {others} (q0)
			    edge              node {word$_j$} (q7)
        (q7) edge [bend left]  node {others} (q0)
			    edge              node {word$_m$} (q8)
        ;
	\node [align =left] [below of =q8] (bbb)  {\\where $q_8$ is a final state.};
\end{tikzpicture}
}}
\\
\\
%\par
\subfloat [A Unified Deterministric Finite Automata for both  concepts $c_1$ and $c_2$]
{
\resizebox{\textwidth}{!}%
{
\begin{tikzpicture}[>=stealth',shorten >=1pt,auto,node distance=1.5cm,baseline]
	\node [align =left] (bb) {Suppose \emph{c1} and \emph{c2} share word1. i.e both concepts start with word1.\\ Then the DFA that represents \emph{c1} and \emph{c2} is:};
  \node[initial,state](q0)  at(-3,-2)  {$q_0$};
  \node[state]         (q1) [right of=q0] at(-3,-2) {$q_1$};
  \node[state]         (q2) [right of=q1] at(-2,-1) {$q_2$};
  \node[state]         (q3) [right of=q2] {$q_3$};
  \node[state,accepting]         (q4) [right of=q3] {$q_4$};
  \node[state]         (q6) [below of=q2] {$q_6$};
  \node[state]         (q7) [right of=q6] {$q_7$};
  \node[state,accepting]         (q8) [right of=q7] {$q_8$};

  \path[->] (q0)  edge [loop above] node {others} (q0)
             edge              node {word$_1$} (q1)
        (q1) edge [bend right=-30]  node[above] {others} (q0);
\path[->]  (q1)   edge      node {word$_2$} (q2)
        (q2) edge [bend left=-40]  node[pos=0.6, above] {others} (q0)
			    edge              node {word$_i$} (q3)
        (q3) edge [bend left=-40]  node[above]  {others} (q0)
			    edge              node {word$_n$} (q4)
        ;
\path[->] (q1)  edge node {$word_3$} (q6);
\path[->] (q6)  edge node {$word_j$} (q7);
\path[->] (q7)  edge node {$word_m$} (q8);
\path[->] (q7)  edge [bend right=-50] node[above] {others} (q0);
\path[->] (q6)  edge [bend left] node[pos=.3,above] {others} (q0);
\end{tikzpicture}
}
}
\\
\hline
\end{tabular}
}
\caption{Merging Deterministic Finite Automata for concepts}
\label{DFA1}
%\vspace{-10pt}  
\end{figure}

%\added{ 
DFAs can effectively process natural language text. DFAs have many advantages for language modeling as well as for mass data processing \cite{finite_Beesley}. %}
The module identifies all distinct concepts in the input list. Every word in a concept is a trigger to transfer the control to a specific state in the concept DFA. Fig.~\ref{DFA1}(a) shows an example of a DFA for a concept. Any concept consists of a number of \emph{n} words,  where $1 \leq \emph{n} \leq 5$. A DFA starts in the initial state $q_0$ and each word causes a transition from a state to another state. If a word appears and does not belong to the concept words (others), then a transition to the initial state $q_0$ occurs. %\added{
This module automatically generates a DFA for each concept. It does not require any specific domain rules or lexico-syntactic rules, which in turn, make it applicable to any domain's concept list. %} 

Every DFA has a final state. When a DFA reaches a final state, it means that the DFA identified a concept. In Fig.~\ref{DFA1}(a) the state $q_4$ is the final state for that DFA. In an analog way, Fig.~\ref{DFA1}(b) shows another DFA for another concept.
The state table generator module joins all DFAs and forms the state table. Fig.~\ref{DFA1}(c) shows an example of merging the DFAs of the two concepts, $c_1$ and $c_2$. We assumed that both concepts start with the same first word $word_1$. As a result, we merged the state $q_0$ and the state $q_5$. We repeated this step for all obtained concepts and their synonyms. As a result, we generated the state table. An example of a state table is shown in Table~\ref{t3}.

\begin{table}%[!t]
\begin{minipage}{\textwidth} 
\caption{A sample mini state table}
\label{t3}
\centering
\resizebox {\textwidth}{!}{
\begin{tabular}{|c||l|l|l|l|l|l|l|l|l|l|l|l|l|}
\hline
& \multicolumn{12}{c|}{input}& \\
\hline
State & root & data & file & independence & item & model & types & warehouse & database & application & management & Others & Term ID  \\ \hline 0 & 1 & 2 & 0 & 0 & 0 & 0 & 0 & 0 & 9 & 0 & 0 & 0 & -1  \\ \hline 1 & $\alpha$ %\footnote{ $\alpha$  A final state}
 &  $\alpha$  &  $\alpha$  &  $\alpha$  &  $\alpha$  &  $\alpha$  &  $\alpha$  &  $\alpha$  &  $\alpha$  &  $\alpha$  &  $\alpha$  &  $\alpha$  & 1  \\ \hline 2 &  $\alpha$  &  $\alpha$  & 3 & 4 & 5 & 6 & 7 & 8 &  $\alpha$  &  $\alpha$  &  $\alpha$  &  $\alpha$  & 2  \\ \hline 3 &  $\alpha$  &  $\alpha$  &  $\alpha$  &  $\alpha$  &  $\alpha$  &  $\alpha$  &  $\alpha$  &  $\alpha$  &  $\alpha$  &  $\alpha$  &  $\alpha$  &  $\alpha$  & 3  \\ \hline 4 &  $\alpha$  &  $\alpha$  &  $\alpha$  &  $\alpha$  &  $\alpha$  &  $\alpha$  &  $\alpha$  &  $\alpha$  &  $\alpha$  &  $\alpha$  &  $\alpha$  &  $\alpha$  & 4  \\ \hline 5 &  $\alpha$  &  $\alpha$  &  $\alpha$  &  $\alpha$  &  $\alpha$  &  $\alpha$  &  $\alpha$  &  $\alpha$  &  $\alpha$  &  $\alpha$  &  $\alpha$  &  $\alpha$  & 5  \\ \hline 6 &  $\alpha$  &  $\alpha$  &  $\alpha$  &  $\alpha$  &  $\alpha$  &  $\alpha$  &  $\alpha$  &  $\alpha$  &  $\alpha$  &  $\alpha$  &  $\alpha$  &  $\alpha$  & 6  \\ \hline 7 &  $\alpha$  &  $\alpha$  &  $\alpha$  &  $\alpha$  &  $\alpha$  &  $\alpha$  &  $\alpha$  &  $\alpha$  &  $\alpha$  &  $\alpha$  &  $\alpha$  &  $\alpha$  & 7  \\ \hline 8 &  $\alpha$  &  $\alpha$  &  $\alpha$  &  $\alpha$  &  $\alpha$  &  $\alpha$  &  $\alpha$  &  $\alpha$  &  $\alpha$  &  $\alpha$  &  $\alpha$  &  $\alpha$  & 8  \\ \hline 9 &  $\alpha$  &  $\alpha$  &  $\alpha$  &  $\alpha$  &  $\alpha$  &  $\alpha$  &  $\alpha$  &  $\alpha$  &  $\alpha$  & 10 & 11 &  $\alpha$  & 9  \\ \hline 10 &  $\alpha$  &  $\alpha$  &  $\alpha$  &  $\alpha$  &  $\alpha$  &  $\alpha$  &  $\alpha$  &  $\alpha$  &  $\alpha$  &  $\alpha$  &  $\alpha$  &  $\alpha$  & 10  \\ \hline 11 &  $\alpha$  &  $\alpha$  &  $\alpha$  &  $\alpha$  &  $\alpha$  &  $\alpha$  &  $\alpha$  &  $\alpha$  & 
 $\alpha$  &  $\alpha$  &  $\alpha$  &  $\alpha$  & 11  \\ \hline 
\end{tabular}}
\end{minipage}
%\vspace{-10pt}
\end{table}

%The state table drives the parser module to tokenise the learning contents resources and to identify the subject concepts embedded in those learning contents. An advantage of this approach is portability. 

In the state table, columns correspond to words of the subject concept list and rows correspond to the DFAs states. A cell has three possible state values, which are: a value of $0$ that represents an unexpected word which causes the parser to start again from state 0 ($q_0$), a positive number N, $ 0 \textless N \textless \alpha$ that means a transition to state N, or a value of $\alpha$ that means a final state. If we reach a final state, then we identified a concept. The value in the last column of a given final state (row) represents a term identifier (ID). 

In programming languages, compilers use this approach to parse program codes. However, we brought it in a new paradigm to parse natural language statements. Also, we automated the process of creating the state table to reduce any configuration complexity or human interaction with the system. This representation allows us to parse all words in a paragraph and to use phrases to index a paragraph. A paragraph may contain one or more concepts. 

By using this approach, portability is achieved since the state table for a subject ontology is constructed automatically. As a result, the knowledge resources can be changed for a different subject and the ontology can be obtained (following the steps in the next subsections) with no extra configuration efforts as the state table is used regardless of the concepts it represents. 
Consequently, developing a new subject ontology requires only changing the learning contents resources.\\

%************************************************************************************************************************************************
\subsubsection{DFA Builder Implementation}

We used the list of frequent concepts and their synonyms as input to build the state table through the use of the deterministic finite automata (DFA) structure, as explained in Section~\ref{dfa}.

To illustrate this step we refer to the concepts in Table~\ref{t2}, Section~\ref{termExtractionImplem}. For simplicity, we omitted the synonyms of these terms. We built a DFA for every concept and obtained the state table, of which an excerpt is shown in Table~\ref{t3}, Section~\ref{dfa}. This state table is used to parse the paragraphs in the textual resources, as well as the questions. Example~\ref{ex1} illustrates the process of parsing a natural language statement using the state table.

\begin{Ex}{Parsing a statement using the state table\\}
\label{ex1}
If we have the following statement ``in database, a data model is ...'' then this statement is parsed and checked against the state table. 

\noindent \emph{Input}:``in database, a data model is ...''\\
\emph{Tokens}: [in, database, a, data, model]\\
\emph{state\_table}: Table~\ref{t3}, Section~\ref{dfa} -- based on the concepts in Table~\ref{t2}.\\
\emph{State}: is the current DFA state. Initially state=0. The first column in the state table holds the state values.\\
\emph{Steps}:
\begin{itemize}
\item The first token is ``in''. We look for its value in the $state\_table[state=0, ``in"]$; as the word ``in'' is not a column in the \emph{state\_table}, the value is taken from the column ``others'' (see also Fig.~\ref{DFA1}).
Consequently, for the word ``in'', the value of $state\_table[state=0,``others"]$ is 0, which means that this word is ignored and the parsing starts again from state 0 with the next token;

\item The next token is ``database'', for which the value of $state\_table [state=0, ``database"]$ is 9, which means go to state 9. The next token is ``data'', and thus,  we find $state\_table[9,``data"] = Acc$ indicating that a final state was reached. Reaching a final state denotes that a term was found, which can be identified from the Term ID (last column in Table~\ref{t3}); in this example the term ID is 9, which can be found in Table~\ref{t2} to be ``database'';

\item We continue until the end of the statement. The result of this step is that we  identified all term IDs which are mentioned in the natural language statement. 
\end{itemize}
\end{Ex}

Through the process mentioned above, another term with the ID 6 is identified, which corresponds to ``data model'' in Table~\ref{t2}. Thus, for the example above, two terms were identified.

We used this state table to parse the course learning resources to identify the subject concepts and to create the transaction DB for the FP-Tree module.

%************************************************************************************************************************
\subsubsection{Transaction database construction approach}
%\noindent \textit{\textbf{Transactions database construction\\}}

%\paragraph{Transactions DB construction}
The state table underpins the parsing module to discover all concepts in a paragraph. The discovered concepts in each paragraph represent a transaction. %for the FP-Tree algorithm. 
We add this transaction to the transaction database (DB). %(for simplification we assigned an ID for every term/concept). 
Algorithm~\ref{A2} shows the pseudo code for extracting the transactions DB and Table~\ref{TDB} displays a subset of these transactions. 

\begin{algorithm}
\caption {Generating the transaction database}
\label {A2}
\begin{algorithmic}[1]
\Procedure{TransactionsDB}{corpus, Concepts, Transactions}
\State $\textit {paragraphs[]} \gets  Split Corpus(Corpus)$
\State $\textit {Transactions} \gets {null}$
\For {\texttt{(p=0, p\textless paragraphs.length(),p++)}}
 \State $\textit{Transactions[p]} \gets \textit{get all concepts(p)} $
\EndFor
\EndProcedure
\end{algorithmic}
\end{algorithm}

In an analog way, the state table is used to parse the user questions in the system-answering system -- this is discussed further in Section~\ref{QASystem}. \\

\begin{table}
\caption{A transaction database sample}
\label{TDB}
\centering
%\begin{tabular}{|p{0.2\linewidth}| p{0.6\linewidth}|}
\begin{tabular}{|c|l|}
 \hline ID & Transaction  \\ \hline 1 & C1, C2, C3  \\ \hline 2 & C2, C4, C5  \\ \hline 3 & C1, C2, C4  \\ \hline 4 & C1,C4  \\ \hline 5 & C1, C3  \\ \hline 
\end{tabular}
\end{table}

\subsubsection{Transaction database construction implementation}

To create the transaction DB, the corpus was divided into paragraphs using the ``openNLP'' library and the co-references were resolved by using Stanford ``coreNLP'' library. 

Each paragraph was parsed using the state table. As a result, each paragraph will add a transaction to the transaction DB. A transaction contains all term IDs which appeared in that paragraph. As a result, we obtain the transaction DB. \\

%******************************************************************************************************************
\subsubsection{FP-Tree construction approach}

%\noindent \textit{\textbf{FP-Tree construction\\}}

\begin{figure}[!h]
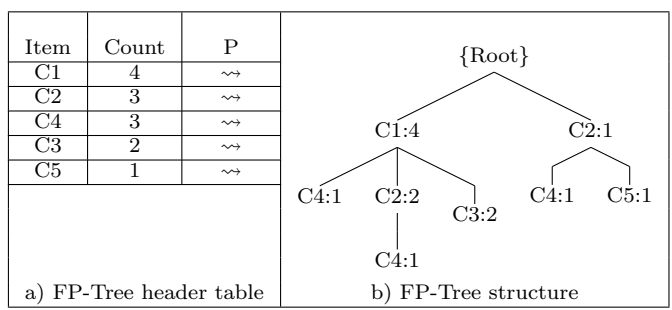

\centering
%\resizebox{\columnwidth}{!}
{
\begin{tabular}{|c|c|c|c|}
\hline
 &   &  &    \\
 %Item & Count & P    & \\ \cline{1-3} 
 % &  &  &  \multirow{7}{*}{\Tree [.\{Root\}  [.C1:4 [.C4:1 ] [.C2:2 [C4:1 ] ] [C3:2 ]  ][.C2:1 [C4:1 ] [C5:1 ]]  ]} \\
Item & Count & P    & \multirow{10}{*}{\Tree [.\{Root\}  [.C1:4 [.C4:1 ] [.C2:2 [C4:1 ] ] [C3:2 ]  ][.C2:1 [C4:1 ] [C5:1 ]]  ]} \\ \cline{1-3} 
C1 & 4 & $\rightsquigarrow$    & \\ \cline{1-3} 
C2 & 3 & $\rightsquigarrow$   & \\ \cline{1-3} 
C4 & 3 & $\rightsquigarrow$   &  \\ \cline{1-3} 
C3 & 2 & $\rightsquigarrow$   & \\ \cline{1-3} 
C5 & 1 & $\rightsquigarrow$   & \\ \cline{1-3} 
\multicolumn{3}{|l|}{ }   & \\
\multicolumn{3}{|l|}{ }   & \\
%\multicolumn{3}{|l|}{ a) FP-Tree Header Table}   &\\
\multicolumn{3}{|l|}{ }  & \\
\multicolumn{3}{|l|}{ }   & \\
% \multicolumn{3}{|l|}{ }  &  b) FP-Tree Structure  \\
\multicolumn{3}{|l|}{ a) FP-Tree header table}  & b) FP-Tree structure \\
\hline
\end{tabular}
}
\caption{FP-Tree construction}
\label{FP}
\end{figure}

The FP-Tree algorithm takes the transaction database as input to generate the FP-Tree structure shown in Fig.~\ref{FP}(b).

\setlength{\textfloatsep}{1pt}
\begin{algorithm}[h]
\caption {FP-Tree construction algorithm}
\label{A11}
\begin{algorithmic}[1]
\Procedure{BuildFPTree}{DB, $ \theta $}
\For {\texttt{(i=0, i $ \textless $ DB.length(),i++)}}
\For {\texttt{(j=0,j $\textless T_i.length()$, j++)}} 
\State Frequency(cj)++
\EndFor

\For {\texttt{(k=0,k $\textless C.length()$, k++)}} 
\If { Frequency($c_k$) $\textgreater \theta$ }
\State FrerquentTerms $\gets$ $c_k$
\EndIf
\EndFor
\EndFor
\State \texttt {L[] $\gets$ {Sort (Frequent Concepts, DES)}}
\State root $\gets$ new node(FP-Tree)
\For {\texttt{(i=0, i $ \textless $ DB.length(),i++)}}
\State Select frequent concepts $\in t_i$
\State Sort ($t_i$) based on L
\State current\_node $\gets$ root
\For {\texttt{(t=0;t$\textless t_i.length()$)}}
\If { \texttt {$c_t \in$ current\_nod.childern}}
\State current\_node.child($c_t$).count ++
\State current\_node $\gets$ current\_node$\shortrightarrow$child($c_t$)
\Else 
\State new current\_node($c_t$)
\State current\_node($c_t$).count =1; 
\State current\_node $\gets$ current\_node$\shortrightarrow$child($c_t$)
\EndIf
\EndFor
\EndFor
\Statex {Description}
\Statex {DB: The transaction database} % contains all transactions in the corpus}
\Statex {T$_i$: A transaction in DB} % consists of one or more concepts}                                     
\Statex {C: A set of all concepts (items).												}
\Statex {c$_i$: A concept in C.                                          }
\Statex {L: Header table contains all concepts sorted according to the concept frequency in descending (DES) order.}
\EndProcedure
\end{algorithmic}

\setlength{\textfloatsep}{0pt}
\end{algorithm}

\setlength{\textfloatsep}{6pt}

A header table is constructed which contains all the items in the transaction DB with their corresponding frequency (count). It also contains a pointer to the first occurrence of an item (concept) in the tree. Thus, every node in the tree has a pointer to the next node occurrence in the tree. By applying the FP-Tree algorithm~\cite{Han_MiningFP} illustrated in Algorithm~\ref{A11}, the FP-Tree structure in Fig.~\ref{FP}(b) is obtained, where every node in the tree corresponds to a concept and its frequency count. \\

%\setlength{\intextsep}{0pt} 

%******************************************************************************************************************
\subsubsection{FP-Tree construction implementation}\label{FPimplem}

%In the next step, t
The FP-Tree algorithm (see Algorithm~\ref{A11}) was used to build the FP-Tree structure. The algorithm gives as an output term-term association values. %\added{
The association value for any two concepts $X$ and $Y$ is obtained by finding the confidence of the rule $X \implies Y$; the rule  $X \implies Y$ holds with confidence $c$ if $c\%$ of the transactions in the database (DB) that contain $X$ also contain $Y$, where DB is the database of all transactions. %} 
These values have been stored in a term-association matrix -- Table~\ref{t5} shows an extract of this matrix.

\begin{table}[!t]
\caption{Term-association matrix sample}
\label{t5}
\centering
\resizebox{\columnwidth}{!}{%
\begin{tabular}{|c|c c c c c|}
\hline
Term & conceptual & data & data model & database & database application\\ \hline data & 2.64151 &  &  &  & \\  data model & 1.50943 & 6.03774 &  &  & \\  database & 2.26415 & 10.9434 & 4.5283 &  &   \\ database application & 0 & 4.15094 & 1.50943 & 4.5283 &   \\  database design & 0 & 1.50943 & 1.50943 & 1.50943 & 0   \\ \hline 
\end{tabular}
}
\vspace{10pt}
\end{table}

%\vspace{5pt}
%******************************************************************************************************************
\subsubsection{FP-Tree customisation approach}
%\noindent \textit{\textbf{FP-Tree customisation\\}}

An item (concept) may appear many times in the original FP-Tree structure. However, in the ontology structure, any concept should appear only once in the concept hierarchy. As a result, multiple occurrences of a concept should be removed. To fulfil this ontology structure requirement, we customised the FP-Tree structure by merging multiple concepts into one instance. 

\begin{figure}
\subfloat[Intial FP-Tree structure \label{FP_a}]{%
\includegraphics[width=0.45\columnwidth] {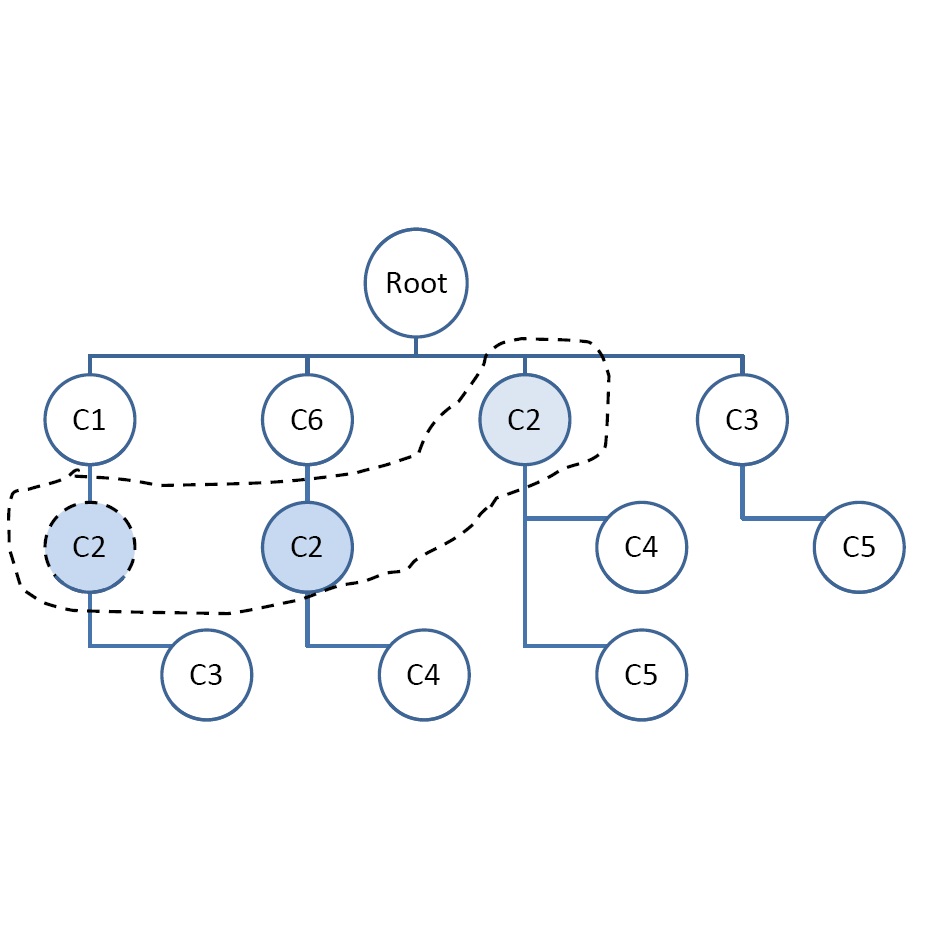}
}
\subfloat[The resulting FP-Tree structure after merging C2 \label{fp_b}]{%
\includegraphics[width=0.45\columnwidth] {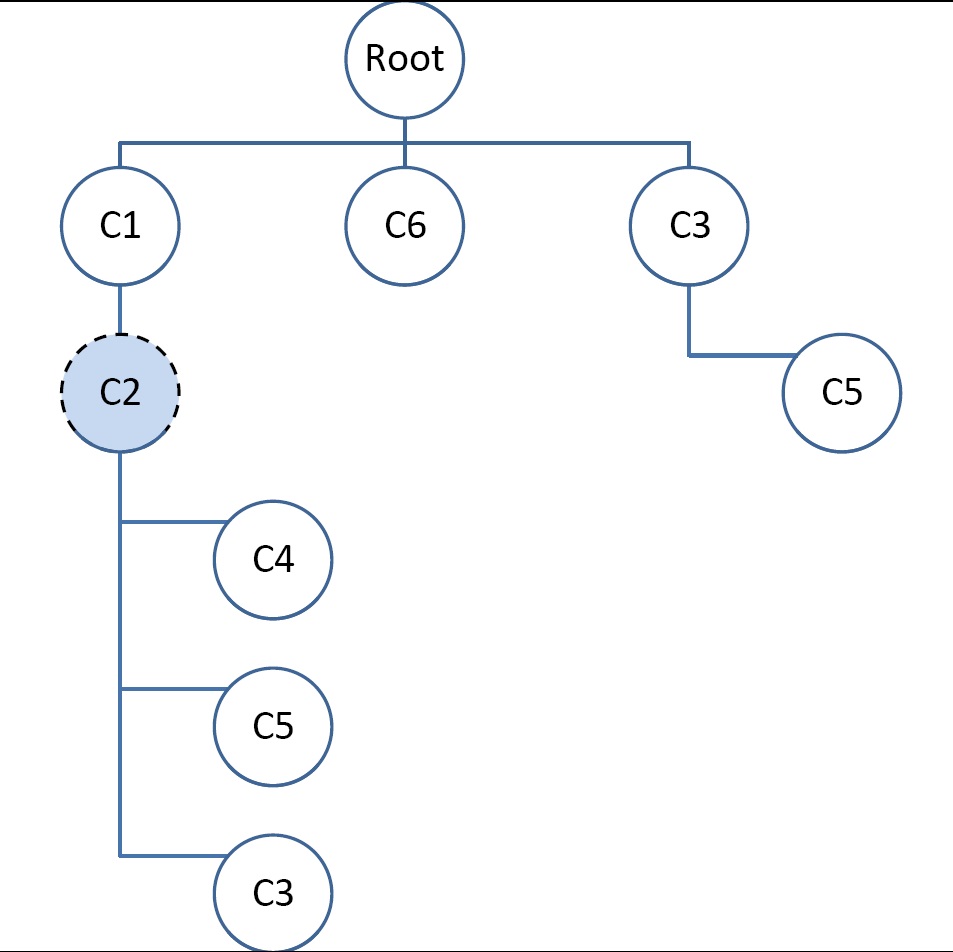}
}
\caption {FP-Tree structure customisation}
\label{fig1_b}
%\vspace{-15pt}
\end{figure}

The criterion used for merging concepts is their frequency. All concepts will be merged under the concept's instance that has the maximum frequency. A top down approach was followed in merging these concepts, by parsing the tree starting from the concept with the highest frequency (top level) down to the lowest frequencies (leaves). All descendant concepts are merged under the concept at the highest level in the structure. %\added{
Starting from the FP-tree's root node, we scan the FP-tree level by level using Breadth-first Search (BFS). For each node (concept), we move all occurrences' subtrees of that concept, if any, under the occurrence that has the maximum frequency count. For example Figure \ref{FP_a}(a) shows an example of an initial FP-tree structure. Let us assume that we are traversing C2 node under the root node, which has two other occurrences in the tree. Suppose the node C2 under the node C1 has the maximum frequency count; we move all C2's subtrees under that node, which results in the structure in Fig.~\ref{fp_b}(b). We repeat this step until there are no more nodes to be merged.
%}

\begin{algorithm}[t]
\caption {Resolving multiple occurrences and the siblings problem}
\label {A3}
\begin{algorithmic}[1]
\Procedure{ConceptsHierarchy}{Transactions}
\State $\textit {FPTree} \gets \textit{Build FP-Tree(Transactions[])}$
\For {\texttt{$c \in Concepts $}}
	\State $\textit{SourceNode}\gets \textit{c}$
	\For {\texttt{$node \in Nodes(c)$}}
		\State $\textit{Merge(SourceNode,c)}$
		\For {\texttt{$child \in child(c)$}}
			\State $\textit{Parent(child)} \gets \textit{SourceNode}$
		\EndFor
	\EndFor
\EndFor
	\For {\texttt{$node \in Nodes(c)$}}
		\State $\textit{A} \gets \textit{Association(c, Parent(c)}$
		\State $\textit{B} \gets \textit{Association(c, GrandParent(c)}$
		\If {$A \textless B$}
			 \State $ \textit {Parent(c)} \gets \textit{GrandParent(c)}$
		\EndIf
	\EndFor
\EndProcedure
\end{algorithmic}
\end{algorithm}

The merging process may generate a hierarchy where sibling concepts appear in a parent-child hierarchy, i.e. concepts may be pushed down to the lower levels in the concept hierarchy. To overcome this problem, we apply a heuristic function to determine if a concept should be moved to become a sibling of another concept. The decision is based on the term-association matrix, which is obtained by transforming the output of the FP-Tree algorithm in a symmetric matrix form, where the rows and columns are concepts, and the cell values represent the associations values (confidence) between concepts.
Table~\ref{t5} in Section~\ref{FPimplem} illustrated a sample of the term-association matrix. %We introduced a complete example in Section~\ref{FPimplem}.

If the association value between the current node and its parent is lower than the association value between the current node and its grandparent, the current node is promoted one level up in the concept hierarchy. Consequently, the current node and its original parent become siblings in the hierarchy. An example of this process in given below. %in Section~\ref{results}.

%\vspace{10pt}

Algorithm~\ref{A3} shows how the FP-Tree structure is refined to solve multiple occurrences of concepts and the siblings problem. A top-down traversal is used for the first aspect, while a bottom-up traversal is used for the second one.  %As a result, the concept hierarchy structure is refined. Based on our validation results, we can claim that the generated concept hierarchy structure is reliable and represents the underlying subject.

%The proposed system  can be generalised for any subject resources. %As a result, the complexity of processing subject resources (learning objects) is reduced. 
%The entire process of generating a subject ontology is illustrated for a particular subject in Section~\ref{results}.
%Figure \ref{f1} shows a subset of the subject course ontology structure. The concepts are organised in a tree structure. 

%******************************************************************************************************************
\subsubsection{FP-Tree customisation implementation}

\begin{figure}[!b]

\centering
\resizebox{\columnwidth}{!}{
\fbox{
\includegraphics{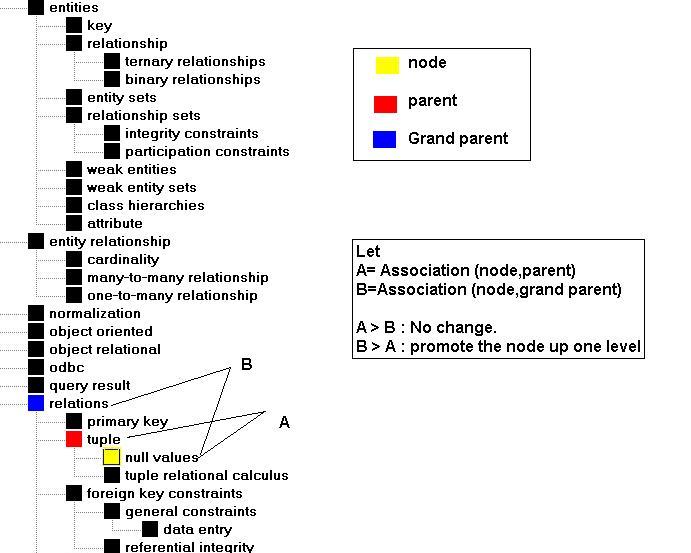}
}
}
\caption{Sample of the concept hierarchy}
\label{f1}
%\vspace{-20pt}
\end{figure}

We used the generated FP-Tree structure and the term-association matrix to enhance the quality of the concept hierarchy by merging co-occurrent concepts in the tree structure and by solving the siblings problem. As a result, the concept hierarchy is obtained -- Fig.~\ref{f1} shows part of the obtained concept hierarchy, as well as the heuristic function used to sort the siblings problem.

%**************************************************************************************************************************************************************************************************************************************************************************************************************************************************************
\subsection{Ontology representation}
As aforementioned, most of the relations among subject concepts are in the form of ``is--a'' relationship or its inverse ``has subtype'' relationship \cite{Boyce_Education}. As a result, after building the concept hierarchy (taxonomy) for the subject we represent relations among concepts in the resulting concept hierarchy using ``is--a'' relationship. Also, in the education-content space, four types of properties were suggested, which are: definition, synonyms, example, and further explanation~\cite{Boyce_Education}. We extended these properties to represent in more detail the underlying subject knowledge by adding the following properties: purpose, syntax, characteristic, advantage, and disadvantage. We used Wordnet synonyms to syntactically extend the property list. We attached these properties to the subject concepts. 

We represent the resulting concept hierarchy and its properties using OWL syntax. Typically, an ontology contains rules and axioms for the underlying domain. However, subject ontologies aim to make a subject knowledge explicit and the application (agent) that utilises the subject ontology can define its own rules to achieve its objectives. In this research, we use the resulting subject ontology to support a question answering system for educational purposes in general and we introduce the question answering system in MOOCs settings. We add rules for the question answering system to achieve its functions. The following section explains how we customise the resulting ontology to support the question answering system.

The final step was the formal representation. The OWL syntax was used to formally represent the subject ontology (concept, property, feedback) triples as illustrated in Fig.~\ref{owl}, and explained in detail in the next section which describes the question-answering system.

%**************************************************************************************************************************************************************************************************************************************************************************************************************************************************************
\section {Phase II: Question-Answering System}\label{QASystem}
%\subsection{Question-Answering System}
In MOOCs settings, thousands of learners enrol in a course and course facilitators are not able to answer all student questions in a timely manner. We propose an approach to automatically answer students' content-related questions by using the subject ontology built as described in the previous section.

\begin{figure}[h]
\resizebox{\columnwidth}{!}{
\frame{
\includegraphics[height=5cm]{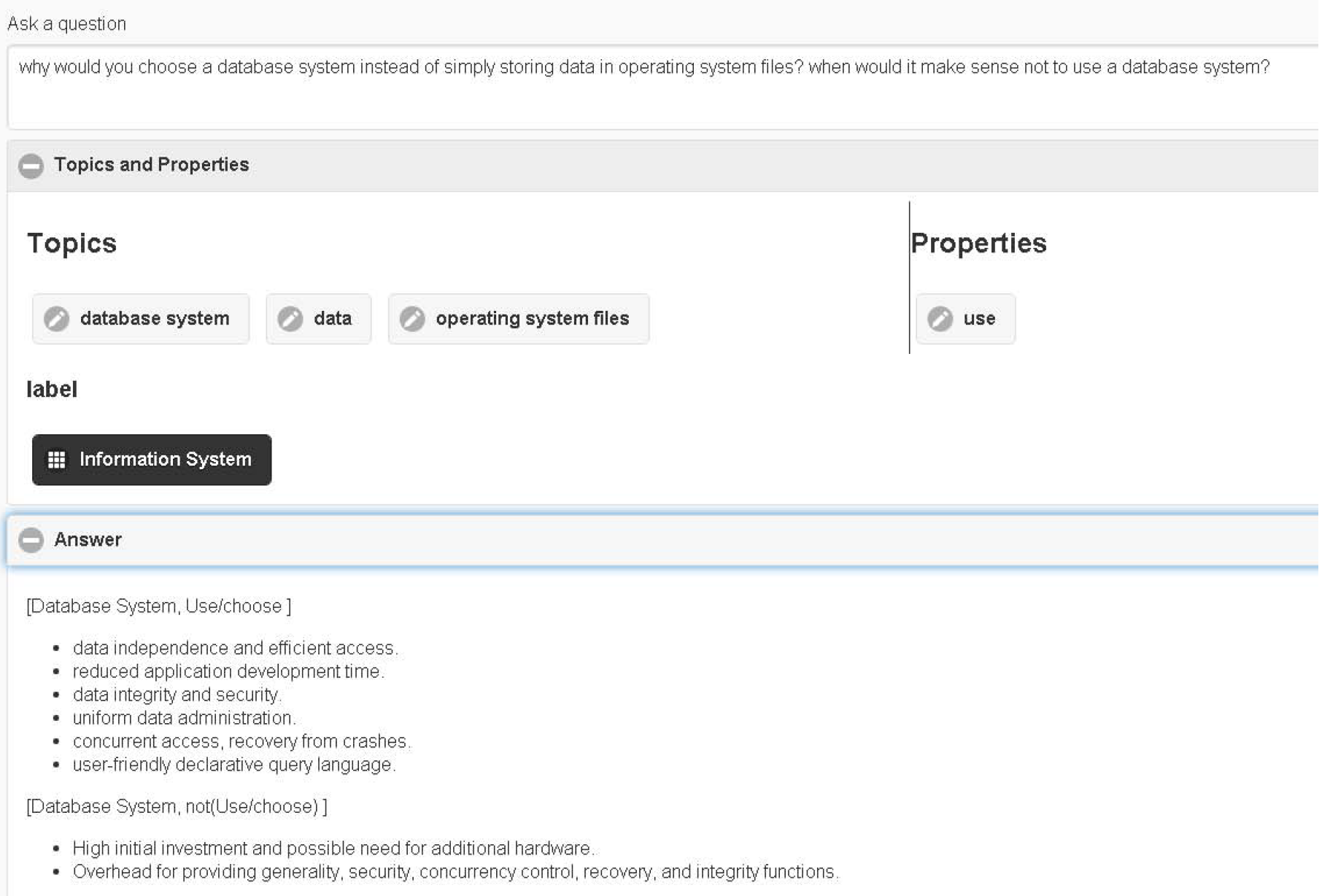}
}
}
\caption{The question-answering system interface}
\label{QA2}
\end{figure}

\begin{figure}%[!t]
\centering
%\resizebox{\textwidth}{!}{
\resizebox{\columnwidth}{!}{
\includegraphics{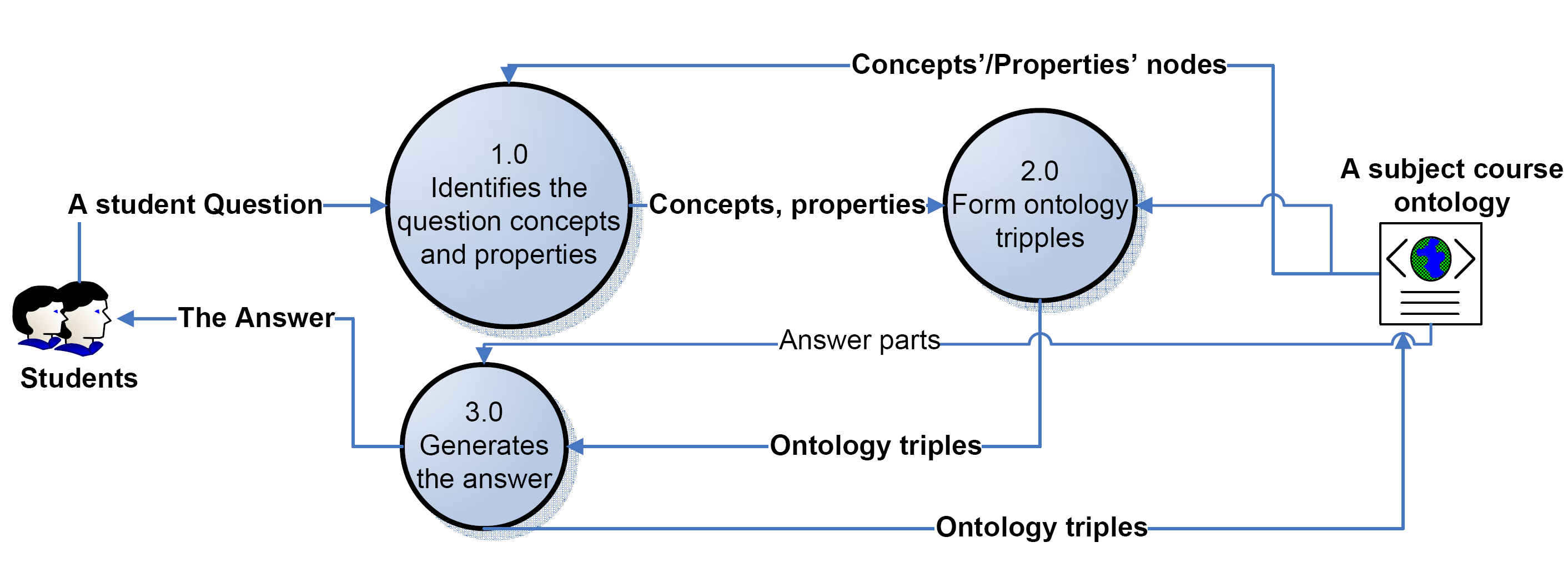}
}
\caption{The question-answering system architecture}
\label{QA}
\end{figure}

Fig.~\ref{QA2} shows the question-answering interface. It allows users to ask a question in English language via a web form. Fig.~\ref{QA} shows the proposed question-answering system. The system queries the subject ontology to identify the concepts and the properties in the learner questions. Then, it retrieves the related information from the subject ontology to answer the learner questions as explained below.

Typically a knowledge-base repository underpins a question answering system. The subject ontology was configured to serve as a knowledge-base for the proposed question-answering system. We used a list of predefined properties in the configuration process. 

The properties were attached to the concepts in the ontology, and each (concept, property) pair was assigned a corresponding feedback, i.e. answer to a question containing the concept and property. Consequently, the knowledge-base for the answering system is represented as (concept, property, feedback) triples.
Fig.~\ref{owl} is a compact OWL code that represents an example of the ontology triple structure for the ``dbms'' concept and the ``definition'' property.

\begin{figure}[t]
\centering
\begin{lstlisting}
<Class rdf:ID="database" />
<Class rdf:ID="concept" />
<Class rdf:ID="property" />
<Class rdf:ID="DBMS" >
<rdfs:subClassOf rdf:resource="database" />
</Class>
<owl:ObjectProperty owl:name="definition"> 
  <owl:domain owl:class="DBMS" /> 
  <feedback> is a computer software application that interacts with
	the user, other applications, and the database itself to capture 
	and analyze data. 
	</feedback>
</owl:ObjectProperty> 
</rdf:RDF>
\end{lstlisting}
\caption{OWL code sample}
\label{owl}
\end{figure}
\color{black}

First, the system accepts the users' queries in the natural language form. Then, it normalises each query by removing all special characters and converts all letters to the lower case form. The query is split into sentences using ``openNLP'' library and co-references are resolved (i.e. clarifying to which word(s) pronouns such as ``it'' refer to), by using the Standford ``coreNLP'' library. 

The sentences are parsed word by word to identify the key concepts and the existing properties. The state table, which is constructed in the ontology building phase, is used to parse these sentences. As a result, the system translates these topics and their properties into (concept, property) pairs and returns all triples that match the identified pairs from the user query.

Finally, the system displays the answer to the user. If the system could not identify any (concept, property) pair, it prompts the user with a ``no answer'' message.

The answers depend on the subject ontology: the more robust the subject ontology, the more reliable and effective the answers returned by the system. As a result, the accuracy and reliability of the answers generated by the system reflect the reliability of the generated subject ontology, as the question-answering system can only identify concepts that appear in the subject ontology. 

%**************************************************************************************************************************************************************************************************************************************************************************************************************************************************************

%\section {Experimental Work and Results}\label{results}

%*****************************************************************************************************************************************************************************************************************************************

\section{Validation}

To validate the generated ontology we measured the impact of using this ontology on the question-answering system for answering content-related questions. The end of chapters questions from the ``Database Design and Management''~\cite{Database_Connolly} %\footnote{Database Systems: Applicational Approach to Design, Implementation, and Management; 4th Edition.} 
 were used to test the system; the contents of this textbook were intentionally left out when building the ontology. In addition, student questions were also used.

The performance of the question-answering system using the proposed subject ontology was compared with the performance of the system when using an ontology produced with the Text2Onto tool.

To evaluate the answers given by the question-answering system, we compared them with the answers from the textbook for 98 questions. We also collected 32 student questions from MOOC forums. The system answered 78 questions out of the 98 textbook questions and 26 questions out of the 32 student questions. %However, 20 questions were not answered by the system. 
The system was not able to answer 20 textbook questions and 6 student questions because their subject terms were not represented in the subject ontology. As a result, all missing terms in the subject ontology will result in no answer for any question related to these terms. Table \ref{t4} shows the percentage of answered/not answered questions.

\begin{table}[!h]
\caption{Experimental results summary}
\label{t4}
\centering
\begin{tabular}{|c|c|c|c|c|c|}
\hline 

Questions & Textbook & Students& \% Texbook & \% Student & Overall Percentage    \\ \hline 
Answered & 78 &26 & 79.6\% & 81.25\% & 80\% \\ \hline 
Not Answered & 20 &6& 20.4\% & 18.75\% & 20\%  \\ \hline 
Total & 98 & 32&  \\ \hline 
\end{tabular}
\end{table}

To identify the best metric for assessing the similarity of the answers, 5 subject experts (who taught ``Database design and management'' at university level in 5 different universities) were asked to evaluate the answers to 10 random questions on a scale from 1 (irrelevant/wrong) to 5 (relevant/accurate). Table~\ref{t3_1} shows the summary of the expert evaluations. 

\begin{table}[!h]
\caption{The experts' evaluation summary}
\label{t3_1}
\centering
\begin{tabular}{|c|c|c|c|}
\hline
Question & Mean & SDEV & LSA similarity  \\ \hline Q1 & 4.86 & 0.38 & 0.643  \\ \hline Q2 & 4.00 & 1.15 & 0.633  \\ \hline Q3 & 3.86 & 1.07 & 0.645  \\ \hline Q4 & 3.14 & 1.86 & 0.259  \\ \hline Q5 & 4.00 & 1.15 & 0.484  \\ \hline Q6 & 3.67 & 1.21 & 0.354  \\ \hline Q7 & 3.14 & 2.04 & 0.83  \\ \hline Q8 & 4.50 & 0.55 & 0.623  \\ \hline Q9 & 4.71 & 0.76 & 0.896  \\ \hline Q10 & 4.43 & 1.13 & 0.594  \\ \hline 
\end{tabular}
\end{table}

To identify the best metric for text similarity, we used the following 7 metrics: (1) greedy comparison based on Wordnet introduced by \cite{Measuring_lintean} to measure the semantic similarity between texts, (2) Latent Semantic Analysis (LSA) using TASA corpus, (3) optimal matching using LSA and TASA corpus, (4) greedy paring using LSA and TASA corpus, (5) greedy comparison using Latent Dirichlet Analysis (LDA) and TASA corpus, (6) Corley and Mihalcea comparer (CM comparer) \cite{Measuring_Corley} and (7) bilingual Evaluation Understudy (BLEU) which is an automated method to evaluate machine translation from a language to another which was extended to find the similarity between texts using the ``Semilar'' toolkit \cite{SEMILAR_Rus}.

Generally, greedy methods calculate the similarity score between TextA and TextB by pairing every word in TextA to all words in TextB. Then, a similarity metric is used to find word to word similarity. Finally, it greedily returns the maximum similarity score between TextA and TextB. The optimal comparer methods represent TextA and TextB as a weighted bipartite graph and find a matching from TextA to TextB which has the maximum weight \cite{AComparison_Rus}.
 
In order to determine the most appropriate measure for our system we used the aforementioned text similarity measures to calculate the similarity between an answer returned by our question-answering system and its answer key which is provided by the textbook authors. 

We used the answers evaluated by experts to benchmark these different measures. We removed the extreme values (which have significant standard deviation), i.e. Q7 where the standard deviation is 2.04, and then calculated the Pearson correlation factor as defined in Equation~\ref{eq1}. 
\begin{gather}
\label {eq1}
r={\sum {(x- \bar{X})(y-\bar{Y})} \over \sqrt{\sum {(x-\bar{X})^2(y-\bar{Y})^2}}}
\end{gather}
%\intertext{}
%\begin{tabular}{l}
%\texttt{
where $\bar{X}$ and $\bar{Y}$ are the means of
the data sets X and Y respectively.
%\end{tabular}\nonumber

Table~\ref{tpearson} summarises the Pearson correlation factor between the text similarity measures and the experts evaluation. The best correlation score of 0.81 is achieved by the LSA based similarity metric. Therefore, the LSA based similarity metric was adopted in the validation step to calculate the similarity score between an answer generated by the question-answering system and its corresponding answer key provided by the textbook authors.

\begin{table}[h]
\centering
\caption{Pearson correlations between similarity metrics and experts' evaluation}
\label{tpearson}
\begin{tabular}{|l|r|} 
\hline Similarity Method & Pearson Correlation  \\ \hline Greedy Comparer WNLin  & -0.12  \\ \hline CM Comparer  & -0.02  \\ \hline LSA  & 0.81  \\ \hline Optimum LSA/Tasa  & 0.12  \\ \hline Greedy LDA/Tasa & -0.11  \\ \hline Dependency WordNet Lesk/Tanim & 0.41  \\ \hline BLEU Comparer & 0.06  \\ \hline 
\end{tabular}
\end{table} 

%\hl {Is any of this description of LSA needed? }
Next, we introduce LSA in more details since we adopt it to be the main similarity metric in validating the returned answers. We used the ``Semilar'' system which is a text similarity tool based on Latent Semantic Analysis (LSA) ~\cite{SEMILAR_Rus}. 

LSA processes a matrix to produce three matrices. This matrix is usually a document-term matrix. The column indexes correspond to the documents in a corpus and the row indexes correspond to the terms in these documents ${M}_{i,j}$, $0 \textless i \textless d$, $0 \textless j \textless t$,  $d\textgreater 0$, $t \textgreater 0$, where $d$ is the number of documents in the corpus and $t$ is the number of different terms in the corpus. It uses the singular value decomposition (SVD) technique which is formally defined in Definition~\ref{d4} to decompose ${M}$ into three matrices ${T}$, ${S}$ and ${D}$.

\begin{onto}
\label{d4}
Let ${M}$ be a matrix with  $d \times t $   dimensions then
${M}$ can be divided into 
${M}_{d \times t} ={T}_{t \times n} {S}_{n \times n} {D}_{n \times d}$
such that ${T}$ and ${D}$ are orthonormal columns and $S$ is diagonal. This is called singular value decomposition of ${M}$.
\end{onto}

Usually ${S}$ contains positive values sorted in descending order. SVD allows a simple strategy for an optimal approximation fit using smaller matrices. It uses the maximum $k$ singular values in the matrix ${S}$ and sets the remaining values in the ${S}$ to zero. Accordingly, it selects the first $k$ columns of the matrix ${T}$ and the first $k$ rows of the matrix ${D}$. Then, it represents the matrix ${M}$ using the new augmented matrices as in the following formula:
${M} \approx {M}^\prime = {T}^ \prime _{d\times k} {S}^\prime_{k \times k} {D}^\prime_{k \times t}$.

%\hl{end}

Applying a Singular Value Decomposition (SVD) on the term-document matrix results in an approximation of it using only the largest $k$ singular values of the decomposition. This represents the LSA model, which is used to find the semantic similarity between words. It can be extended to find similarity between documents by aggregating the semantic similarity measures for all words in these documents. 
LSA is an effective tool in detecting word to word similarity beyond the lexical word to word synonyms. LSA is underpinned by the idea that the aggregate of all the word contexts in which a given word does/does not appear provides a set of mutual constraints that largely determines the similarity of meaning between words and sets of words. %to each other.

We used the LSA text similarity tool (``Semilar'') to compare the answer keys of the end of chapter questions to the 78 answers returned by the proposed question-answering system. 
Table~\ref{texp} shows the similarity summary. We divided the table into 10 ranges; for every range, we count the number of answers that fall in that range. 
For the 10 questions evaluated by the experts, the last column in Table \ref{t3_1} shows the LSA based text similarity between the answer key and the automatic generated answer pairs.

\begin{table}[!h]
\caption{LSA-based similarity (answer vs answer key)}
\label{texp}
\centering
\begin{tabular}{|c|c|}
\hline
Range & Count  \\ \hline 0.00 - 0.10 & 2  \\ \hline 0.10 - 0.20 & 0  \\ \hline 0.20 - 0.30 & 2  \\ \hline 0.30 - 0.40 & 2  \\ \hline 0.40 - 0.50 & 1  \\ \hline 0.50 - 0.60 & 2  \\ \hline 0.60 - 0.70 & 14  \\ \hline 0.70 - 0.80 & 2  \\ \hline 0.80 - 0.90 & 12  \\ \hline 0.90 - 1.00 & 41  \\ \hline 
\end{tabular}
\end{table}

There are 71 out of 78 answers (91\%) with a value above 60\% for the LSA metric. Moreover, the majority of the answers (53 answers representing 68\%) have similarity values above 80\%.

A possible reason for having answers which have a low similarity ratio is that these questions ask about multiple concepts and some of these terms were not selected among the subject ontology terms. As a result, the system will answer part of the question and ignore the remaining part of the question. In fact, this occurred for questions 3 and 4 of the ones evaluated by the experts.

Some concepts were not listed in the subject course ontology due to the following reasons:
\begin{itemize}
\item These concepts are not frequent concepts in the corpus used to build the ontology;

\item These concepts are frequent in the corpus, however, they are also frequent in the COCA corpus; as a result, the proposed ontology system will remove these concepts from the ontology concept list;

\item These concepts are synonyms that have not been generated by the ``Wordnet'' synonyms tool. 
\end{itemize}

This drawback can be overcome by allowing course facilitators to add any missing concepts to the concept list. This task does not require any technical experience. Also, since we proposed an automated state table construction module, the following modules in the subject course ontology system do not require any modification. \\

Finally, we used the comparative validation approach~\cite{Zouaq2009} to validate the generated ontology. We ran the Text2Onto tool~\cite{Text2Onto} on the same corpus to generate a subject ontology and used the generated ontology in the question-answering system to answer the 98 texbook questions and the 32 student questions. Table~\ref{Onto_Validation} shows the accuracy (i.e. percentage of answered questions) of the question-answering system using both Text2Onto and our proposed ontology.

\begin{table}
\centering
\caption {Question-answering accuracy using Text2Oonto and the proposed ontology }
\label{Onto_Validation}
\centering
\resizebox{\columnwidth}{!}{%
\begin{tabular}
{|c|c|c|c|c|c|} 
\hline & \multicolumn{2}{c}{Textbook questions} & \multicolumn{2}{|c|}{Student questions} & \\ 
\hline Ontology & Answered & Not Answered & Answered & Not Answered & Accuracy  \\ \hline 
Text2Onto & 28 & 70 & 9&23& 28.4\%  \\ \hline 
Proposed ontology & 78 & 20 & 26 & 6 &80\%  \\ \hline 
\end{tabular}
}
\vspace{10pt}
\end{table}

\begin{figure}[h]
\subfloat[Text2Onto concept hierarchy sample\label{tax_a}]{%
\includegraphics[width=0.47\columnwidth] {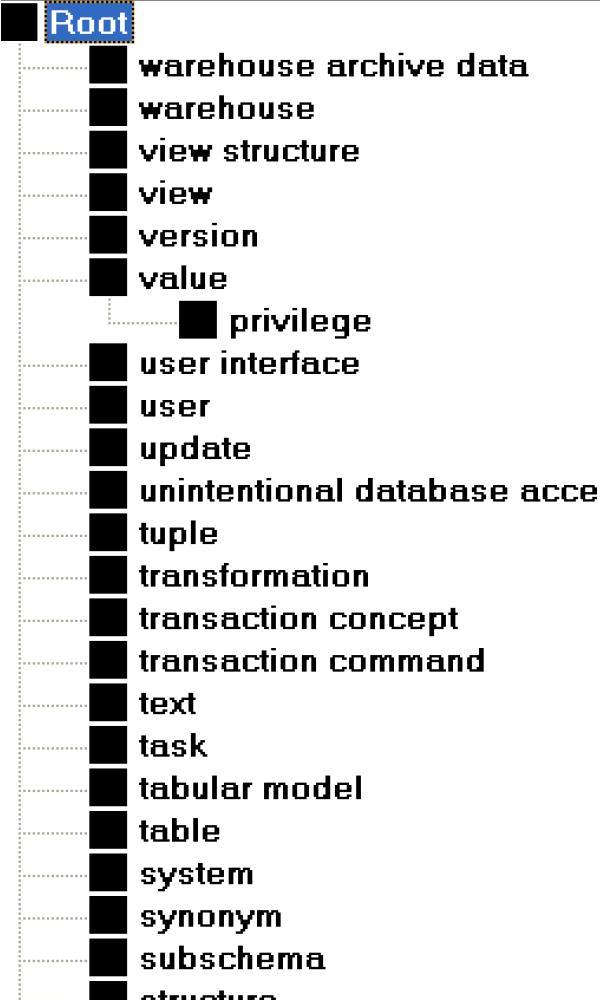}
}
\hspace{10pt}
\subfloat[Proposed concept hierarchy sample\label{tax_b}]{%
\includegraphics[width=0.47\columnwidth] {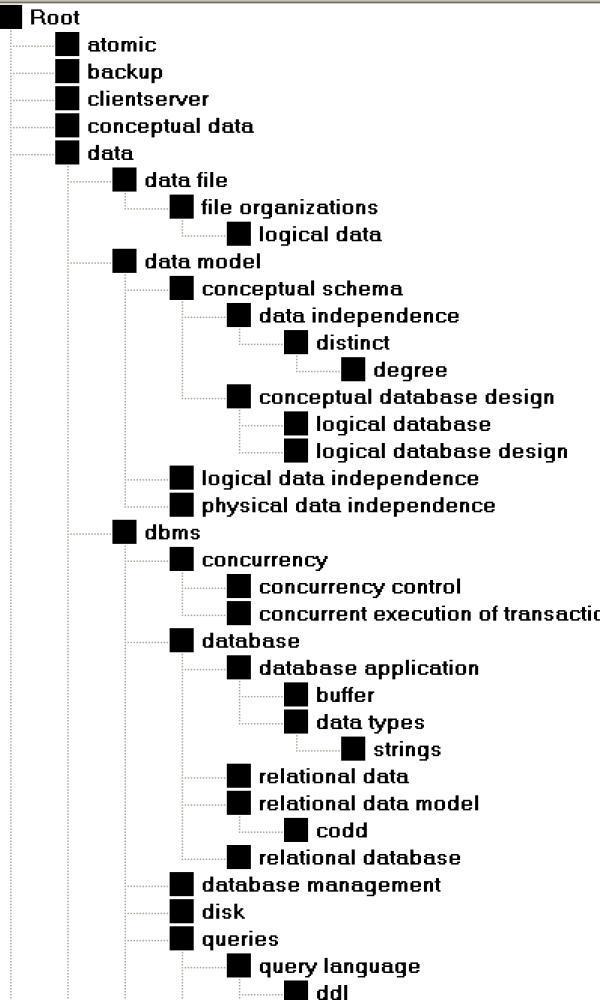}
}
\caption {Concept hierarchies generated by Text2Onto (a) and our ontology learning approach (b)}
\label{fig2_b}
%\vspace{-15pt}
\end{figure}

Our approach outperforms the Text2Onto tool. We noticed that the Text2Onto tool generated a long list of irrelevant terms when compared with our proposed system, which affected the quality of the generated ontology. Moreover, Text2Onto generated a flat concept hierarchy -- most of the concepts appear immediately under the root node as shown in Fig.~\ref{fig2_b}. As a result, the question answering system performed poorly when using this ontology.

%**************************************************************************************************************************************************************************************************************************************************************************************************************************************************************

\section{Summary and Conclusion}

In this research, we proposed a framework to automatically build a subject ontology from overlapping heterogeneous learning contents in plain text format. 

We represented the subject terms and concepts using the Deterministic Finite Automata (DFA) notation. We developed a module that takes the subject concepts and generates a state table for these concepts. This state table is used in the following modules to detect the subject concepts for the concept hierarchy construction and for parsing the questions in the question-answering system. 

We proposed a novel data mining-based technique to construct the concept hierarchy for the identified concepts. A heuristic function based on concept association mining drives the concept hierarchy-construction module to enhance the quality of the concept hierarchy structure by resolving multiple occurrences within the hierarchy and by solving the siblings problem. The DFA representation and the concept-hierarchy construction modules make our approach applicable to different subjects.

Finally, we proposed a question-answering (Q\&A) system underpinned by the resulting subject ontology. The question-answering system answers content-related questions. The system targets the massive open online courses platforms to fulfil the cognitive needs of MOOCs registrants. 
The proposed ontology learning systems is suitable for e-learning environments, especially for MOOCs settings and educators with novice IT skills.%\hl{refine}

We validated the proposed systems using the subject course experts and using an LSA based text similarity metric. We used a set of content related questions from a ``Database Design and Management'' textbook, as well as 32 student related questions, to test the question-answering system and evaluated the quality and the correctness of the returned answers. The results support our hypothesis, as the system was able to correctly answer 80\% of the questions, which is significantly more than the 28.4\% obtained when using Text2Onto.

%\hl{limitation of missing terms (questions 3, 4 and 6) - discuss further; discuss the complexity of questions, e.g. question 4 asked to compare different concepts; }.

A limitation of our approach occurs when the system fails to capture some concepts of the underlying subject. It propagates to the question-answering module, where missing concepts are also not captured in students' questions. As a result, it will generate an incomplete answer (a partial answer) to that question. This explains the low similarity values in Table~\ref{t3_1} for the questions 3 and 6. Also, the complexity of the questions may affect the quality of the generated answers. Questions at the higher levels of Bloom's taxonomy may not be answered correctly as occurred in question 7 in Table~\ref{t3_1}. %\hl{
However, we can overcome this limitation by initiating a dialogue with the learner to ask them to split their questions into multiple sentences.%}

%\hl{Conclusion - generality of approach and its use for other personalisation opportunities in MOOCs.}
There are many opportunities to use the proposed system in MOOCs. The resulting subject ontology can support pedagogical agents to support both collaborative and individualised learning, as well as the students' cognitive processes. On the other hand, the question-answering system can be extended to analyse students cognitive needs and give feedback for course facilitators about students learning. 

%\hl{
We intend to extend our research to allow instructors or even learners to edit the subject ontology part. Ontology editing has a two-fold value. First, it allows instructors/learners to add any missing concepts that have not been captured by the proposed system; as a result, it enhances the quality of the subject ontology which in turn improves the accuracy of the question-answering system. Second, it builds consensus for the subject ontology which is an important part of the ontology definition and cannot be achieved without having multiple perspectives reflected in the subject ontology. %}

\vspace{5mm}

\noindent \textbf{Funding}: This study was self-funded as a part of a completed PhD by \emph{Safwan Shatnawi}, supervised by \emph{Mohamed Medhat Gaber} and \emph{Mihaela Cocea}.

\noindent \textbf{Conflict of Interest}: The authors declare that they have no conflict of interest.

\bibliographystyle{spmpsci}      % mathematics and physical sciences
\bibliography{bib}   % name your BibTeX data base

\begin{thebibliography}{10}
\providecommand{\url}[1]{{#1}}
\providecommand{\urlprefix}{URL }
\expandafter\ifx\csname urlstyle\endcsname\relax
  \providecommand{\doi}[1]{DOI~\discretionary{}{}{}#1}\else
  \providecommand{\doi}{DOI~\discretionary{}{}{}\begingroup
  \urlstyle{rm}\Url}\fi

\bibitem{Rakesh_Toward}
Agrawal, R., Golshan, B., Papalexakis, E.E.: Toward data-driven design of
  educational courses: {A} feasibility study.
\newblock JEDM \textbf{8}(1), 1--21 (2016)

\bibitem{Ontologies_Al_Yahya}
Al-Yahya, M., George, R., Alfaries, A.: Ontologies in e-learning: review of the
  literature.
\newblock International Journal of Software Engineering and Its Applications
  \textbf{9}(2), 67--84 (2015)

\bibitem{Arabshian2012}
Arabshian, K., Danielsen, P.J., Afroz, S.: Lexont: A semi-automatic ontology
  creation tool for programmable web.
\newblock In: AAAI Spring Symposium: Intelligent Web Services Meet Social
  Computing, pp. 2--8 (2012)

\bibitem{Aroyo2006}
Aroyo, L., Denaux, R., Dimitrova, V., Pye, M.: Interactive ontology-based user
  knowledge acquisition: A case study.
\newblock In: ESWC, pp. 560--574 (2006).
\newblock \doi{10.1007/11762256_41}

\bibitem{Aroyo2004}
Aroyo, L., Dicheva, D.: The new challenges for e-learning: The educational
  semantic web.
\newblock Educational Technology \& Society \textbf{7}(4), 59--69 (2004)

\bibitem{Balachandran2016}
Balachandran, K., Ranathunga, S.: Domain-specific term extraction for concept
  identification in ontology construction.
\newblock In: 2016 IEEE/WIC/ACM International Conference on Web Intelligence
  (WI), pp. 34--41 (2016).
\newblock \doi{10.1109/WI.2016.0016}

\bibitem{finite_Beesley}
Beesley, K.R., Karttunen, L.: Finite-state morphology: Xerox tools and
  techniques.
\newblock CSLI, Stanford  (2003)

\bibitem{Berners-Lee2001}
Berners-Lee, T., Hendler, J., Lassila, O.: The semantic web.
\newblock Scientific american \textbf{284}(5), 28--37 (2001)

\bibitem{Boyce_Education}
Boyce, S., Pahl, C.: Developing domain ontologies for course content.
\newblock Educational Technology \& Society \textbf{10}(3), 275--288 (2007)

\bibitem{Buitelaar2005}
Buitelaar, P., Cimiano, P., Magnini, B.: Ontology learning from text: An
  overview.
\newblock In: Ontology learning from text: methods, evaluation and
  applications, pp. 3--12. IOS press (2005)

\bibitem{Adding_chen}
Chen, R.C., Lee, Y.C., Pan, R.H.: Adding new concepts on the domain ontology
  based on semantic similarity.
\newblock In: International Conference on Business and information, pp. 12--14.
  Citeseer (2006)

\bibitem{Chowdhury2010}
Chowdhury, G.: Introduction to modern information retrieval.
\newblock Facet publishing (2010)

\bibitem{Text2Onto}
Cimiano, P., V\"{o}lker, J.: Text2{O}nto: A framework for ontology learning and
  data-driven change discovery.
\newblock In: NLDB, pp. 227--238 (2005)

\bibitem{supporting_cesar}
Coll, C., Rochera, M.J., de~Gispert, I.: Supporting online collaborative
  learning in small groups: teacher feedback on learning content, academic task
  and social participation.
\newblock Computers \& Education \textbf{75}, 53 -- 64 (2014)

\bibitem{Database_Connolly}
Connolly, T.M., Begg, C.: Database Systems: A Practical Approach to Design,
  Implementation, and Management, 3rd edn. (2001)

\bibitem{Measuring_Corley}
Corley, C., Mihalcea, R.: Measuring the semantic similarity of texts.
\newblock In: ACL Workshop on Empirical Modeling of Semantic Equivalence and
  Entailment, pp. 13--18 (2005)

\bibitem{An_Intelleginet_Rebecca}
Crowley, R.S., Medvedeva, O.: An intelligent tutoring system for visual
  classification problem solving.
\newblock Artif. Intell. Med. \textbf{36}(1), 85--117 (2006).
\newblock \doi{10.1016/j.artmed.2005.01.005}

\bibitem{COCA}
Davies, M.: The corpus of contemporary american english: 450 million words,
  1990-present (2008-)

\bibitem{Dellschaft_onHow}
Dellschaft, K., Staab, S.: On how to perform a gold standard based evaluation
  of ontology learning.
\newblock In: International Semantic Web Conference, vol. 4273, pp. 228--241.
  Springer (2006)

\bibitem{TM4L_Dicheva}
Dicheva, D., Dichev, C.: {TM4L}: Creating and browsing educational topic maps.
\newblock British Journal of Educational Technology \textbf{37}(3), 391--404
  (2006).
\newblock \doi{10.1111/j.1467-8535.2006.00612.x}

\bibitem{OntoGain}
Drymonas, E., Zervanou, K., Petrakis, E.G.: Unsupervised ontology acquisition
  from plain texts: The ontogain system.
\newblock In: NLDB, pp. 277--287. Springer (2010)

\bibitem{Elmasri2010}
Elmasri, R., Navathe, S.B.: Fundamentals of database systems.
\newblock Pearson (2010)

\bibitem{espinosa_extasem}
Espinosa-Anke, L., Saggion, H., Ronzano, F., Navigli, R.: Extasem! extending,
  taxonomizing and semantifying domain terminologies.
\newblock In: Proceedings of the 30th Conference on Artificial Intelligence
  (AAAI16) (2016)

\bibitem{tm}
Feinerer, I., Hornik, K.: tm: Text Mining Package (2015)

\bibitem{WordnetR}
Feinerer, I., Hornik, K.: wordnet: WordNet Interface (2015)

\bibitem{AnIntelligent_Feng}
Feng, D., Shaw, E., Kim, J., Hovy, E.: An intelligent discussion-bot for
  answering student queries in threaded discussions.
\newblock In: IUI, pp. 171--177 (2006).
\newblock \doi{10.1145/1111449.1111488}

\bibitem{Automating_Fiedler}
Fiedler, A., Tsovaltzi, D.: Automating hinting in an intelligent tutorial
  dialog system for mathematics.
\newblock IJCAI Workshop on Knowledge Representation and Automated Reasoning
  for E-Learning Systems pp. 23--35 (2003)

\bibitem{Research_Frederick}
Gravetter, F.J., Forzano, L.A.B.: Research Methods for the Behavioral Sciences,
  5th edition edn.
\newblock CENGAGE Learning (2015)

\bibitem{Atranslation_Gruber}
Gruber, T.R.: A translation approach to portable ontology specifications.
\newblock Knowledge Acquisition \textbf{5}, 199--220 (1993)

\bibitem{Gupta2008}
Gupta, S., Mittal, S., Mittal, A.: Eureqa: Overcoming the digital divide
  through a multidocument {QA} system for e-learning.
\newblock In: National conference on emerging trends in Information Technology
  (2008)

\bibitem{Han_MiningFP}
Han, J., Pei, J., Yin, Y.: Mining frequent patterns without candidate
  generation.
\newblock SIGMOD Rec. \textbf{29}(2), 1--12 (2000).
\newblock \doi{10.1145/335191.335372}

\bibitem{Ontology_Extraction_hatala}
Hatala, M., Gasevic, D., Siadaty, M., Jovanovic, J., Torniai, C.: Ontology
  extraction tools: An empirical study with educators.
\newblock Learning Technologies, IEEE T on \textbf{5}(3), 275--289 (2012).
\newblock \doi{10.1109/TLT.2012.9}

\bibitem{HayesCollaborative}
Hayes, P., Eskridge, T.C., Saavedra, R., Reichherzer, T., Mehrotra, M.,
  Bobrovnikoff, D.: Collaborative knowledge capture in ontologies.
\newblock In: K-CAP, pp. 99--106 (2005)

\bibitem{Reasoning_Henze}
Henze, N., Dolog, P., Nejdl, W.: Reasoning and ontologies for personalized
  e-learning in the semantic web.
\newblock Journal of Educational Technology \& Society \textbf{7}(4), 82--97
  (2004)

\bibitem{openNLP}
Hornik, K.: openNLP: Apache OpenNLP Tools Interface (2014)

\bibitem{RWeka}
Hornik, K., Buchta, C., Zeileis, A.: Open-source machine learning: {R} meets
  {Weka}.
\newblock Computational Statistics \textbf{24}(2), 225--232 (2009).
\newblock \doi{10.1007/s00180-008-0119-7}

\bibitem{Asemantic_Seiji}
Isotani, S., Mizoguchi, R., Isotani, S., Capeli, O.M., Isotani, N.,
  de~Albuquerque, A.R.L., Bittencourt, I.I., Jaques, P.: A semantic web-based
  authoring tool to facilitate the planning of collaborative learning scenarios
  compliant with learning theories.
\newblock Computers \& Education \textbf{63}, 267 -- 284 (2013).
\newblock \doi{http://dx.doi.org/10.1016/j.compedu.2012.12.009}

\bibitem{jiang_crctol}
Jiang, X., Tan, A.H.: Crctol: A semantic-based domain ontology learning system.
\newblock Journal of the Association for Information Science and Technology
  \textbf{61}(1), 150--168 (2010)

\bibitem{Kamel2013}
Kamel, M., Aussenac-Gilles, N., Buscaldi, D., Comparot, C.: A semi-automatic
  approach for building ontologies from acollection of structured web
  documents.
\newblock In: K-CAP, pp. 139--140 (2013)

\bibitem{taxofinder_kang}
Kang, Y.B., Haghigh, P.D., Burstein, F.: Taxofinder: A graph-based approach for
  taxonomy learning.
\newblock IEEE Transactions on Knowledge and Data Engineering \textbf{28}(2),
  524--536 (2016)

\bibitem{education_kasimati}
Kasimati, A., Zamani, E.: Education and learning in the semantic web.
\newblock In: 15th Panhellenic Conference on Informatics, pp. 338--344 (2011).
\newblock \doi{10.1109/PCI.2011.40}

\bibitem{start_katz}
Katz, B., Borchardt, G.C., Felshin, S.: Natural language annotations for
  question answering.
\newblock In: FLAIRS, pp. 303 -- 306 (1993)

\bibitem{Leveraging_Kazi}
Kazi, H., Haddawy, P., Suebnukarn, S.: Leveraging a domain ontology to increase
  the quality of feedback in an intelligent tutoring system.
\newblock In: ITS Proceedings, pp. 75--84 (2010).
\newblock \doi{10.1007/978-3-642-13388-6_12}

\bibitem{Measuring_Jianhua}
Li, J., Wang, X.: To discover and integrate the education resources based on
  semantic web.
\newblock In: ICMTMA, pp. 1264--1267 (2013).
\newblock \doi{10.1109/ICMTMA.2013.311}

\bibitem{Measuring_lintean}
Lintean, M., Rus, V.: Measuring semantic similarity in short texts through
  greedy pairing and word semantics.
\newblock In: FLAIRS, pp. 13--18 (2012)

\bibitem{Kate_Ontology}
Litherland, K., Carmichael, P., MartÃ­nez-GarcÃ­a, A.: Ontology-based
  e-assessment for accounting: Outcomes of a pilot study and future prospects.
\newblock Journal of Accounting Education \textbf{31}(2), 162 -- 176 (2013)

\bibitem{CoreNLP}
Manning, C.D., Surdeanu, M., Bauer, J., Finkel, J., Bethard, S.J., McClosky,
  D.: The {Stanford} {CoreNLP} natural language processing toolkit.
\newblock In: ACL, pp. 55--60 (2014)

\bibitem{NLP_Maynard}
Maynard, D., Li, Y., Peters, W.: {NLP} techniques for term extraction and
  ontology population.
\newblock In: Conference on Ontology Learning and Population: Bridging the Gap
  Between Text and Knowledge, pp. 107--127 (2008)

\bibitem{semantic_Meijer}
Meijer, K., Frasincar, F., Hogenboom, F.: A semantic approach for extracting
  domain taxonomies from text.
\newblock Decision Support Systems \textbf{62}, 78--93 (2014)

\bibitem{Question_Moll}
Moll{\'a}, D., Vicedo, J.L.: Question answering in restricted domains: An
  overview.
\newblock Computational Linguistics \textbf{33}(1), 41--61 (2007)

\bibitem{Pedro_An_ontology}
Munoz-Merino, P.J., Pardo, A., Scheffel, M., Niemann, K., Wolpers, M., Leony,
  D., Kloos, C.D.: An ontological framework for adaptive feedback to support
  students while programming.
\newblock In: ISWC (2011)

\bibitem{enterprise_nayak}
Nayak, A., Agarwal, J., Yadav, V., Pasha, S.: Enterprise architecture for
  semantic web mining in education.
\newblock In: ICCEE, pp. 23--26 (2009).
\newblock \doi{10.1109/ICCEE.2009.222}

\bibitem{Mining_Viral}
Parekh, V., Gwo, J.P., Finin., T.W.: Mining domain specific texts and
  glossaries to evaluate and enrich domain ontologies.
\newblock In: IKE, pp. 485--490 (2004)

\bibitem{Poon2010}
Poon, H., Domingos, P.: Unsupervised ontology induction from text.
\newblock In: Proceedings of the 48th annual meeting of the Association for
  Computational Linguistics, pp. 296--305. Association for Computational
  Linguistics (2010)

\bibitem{Ramakrishnan2000}
Ramakrishnan, R., Gehrke, J.: Database management systems.
\newblock McGraw Hill (2000)

\bibitem{AComparison_Rus}
Rus, V., Lintean, M.: A comparison of greedy and optimal assessment of natural
  language student input using word-to-word similarity metrics.
\newblock In: Workshop on Building Educational Applications Using {NLP}, pp.
  157--162 (2012)

\bibitem{SEMILAR_Rus}
Rus, V., Lintean, M., Banjade, R., Niraula, N., Stefanescu, D.: {SEMILAR}: The
  semantic similarity toolkit.
\newblock In: ACL, pp. 163--168 (2013)

\bibitem{IDEAL_14}
Shatnawi, S., Gaber, M.M., Cocea, M.: Automatic content related feedback for
  moocs based on course domain ontology.
\newblock In: IDEAL, Lecture Notes in Computer Science. Springer (2014)

\bibitem{sowa_knowledge}
Sowa, J.F., et~al.: Knowledge representation: logical, philosophical, and
  computational foundations, vol.~13.
\newblock MIT Press (2000)

\bibitem{staab_Handbook_on_ontology}
Studer, R., Staab, S.: Handbook on Ontologies.
\newblock International Handbooks on Information Systems. Springer (2009)

\bibitem{Anapproach_Valencia}
Valencia-Garc{\'i}a, R., Castellanos~Nieves, D., Vivancos~Vicente, P.J.,
  Fern{\'a}ndez~Breis, J.T., Mart{\'i}nez-B{\'e}jar, R.,
  Garc{\'i}a~S{\'a}nchez, F.: An approach for ontology building from text
  supported by {NLP} techniques.
\newblock In: Current Topics in Artificial Intelligence, pp. 126--135. Springer
  Berlin Heidelberg (2004).
\newblock \doi{10.1007/978-3-540-25945-9_13}

\bibitem{velardi_ontolearn}
Velardi, P., Faralli, S., Navigli, R.: Ontolearn reloaded: A graph-based
  algorithm for taxonomy induction.
\newblock Computational Linguistics \textbf{39}(3), 665--707 (2013)

\bibitem{Velardi2005}
Velardi, P., Navigli, R., Neri, R.: Evaluation of ontolearn, a methodology for
  automatic learning of domain ontologies.
\newblock In: Ontology learning from text: Methods, applications and
  evaluation, pp. 92--106 (2005)

\bibitem{abstract_verma}
Verma, A.: An abstract framework for ontology evaluation.
\newblock In: ICDSE, pp. 1--6. IEEE (2016)

\bibitem{WordnetJ}
Wallace, M.: Jawbone Java WordNet API (2007)

\bibitem{Instructor_Dunwei}
Wen, D., Cuzzola, J., Brown, L., Kinshuk: Instructor-aided asynchronous
  question answering system for online education and distance learning.
\newblock The International Review of Research in Open and Distributed Learning
  \textbf{13}(5), 102--125 (2012)

\bibitem{Wong2007}
Wong, W., Liu, W., Bennamoun, M.: Determining termhood for learning domain
  ontologies using domain prevalence and tendency.
\newblock In: Proceedings of the Sixth Australasian Conference on Data Mining
  and Analytics - Volume 70, AusDM '07, pp. 47--54. Australian Computer
  Society, Inc., Darlinghurst, Australia, Australia (2007).
\newblock \urlprefix\url{http://dl.acm.org/citation.cfm?id=1378245.1378253}

\bibitem{Yang2004}
Yang, S.J., Chen, I.Y.L., Shao, N.W., et~al.: Ontology enabled annotation and
  knowledge management for collaborative learning in virtual learning
  community.
\newblock Educational Technology \& Society \textbf{7}(4), 70--81 (2004)

\bibitem{Zouaq2011}
Zouaq, A., Gasevic, D., Hatala, M.: Towards open ontology learning and
  filtering.
\newblock Information Systems \textbf{36}(7), 1064--1081 (2011)

\bibitem{Zouaq2009}
Zouaq, A., Nkambou, R.: Evaluating the generation of domain ontologies in the
  knowledge puzzle project.
\newblock Knowledge and Data Engineering, IEEE Trans. on \textbf{21}(11),
  1559--1572 (2009).
\newblock \doi{10.1109/TKDE.2009.25}

\end{thebibliography}

%\begin{thebibliography}{}
% etc
%\end{thebibliography}

\end{document}